\documentclass[journal abbreviation, manuscript]{copernicus}

\usepackage{subfig}
\usepackage{enumitem}
\usepackage{booktabs}
\usepackage{multirow}
\usepackage{array}
\usepackage{graphicx}
\usepackage[acronym]{glossaries}
\usepackage{verbatim}

\usepackage[absolute,overlay]{textpos} % For absolute positioning

\usepackage{caption}
\usepackage{afterpage} % Add this to handle page breaks more reliably
\usepackage{float}

\usepackage{todonotes}

\usepackage{subcaption}

\makeglossaries

\newacronym{RoI}{RoI}{Region of Interest}
\newacronym{OSM}{OSM}{OpenStreetMap}
\newacronym{PSR}{PSR}{PlanetScope Surface Reflectance}

\graphicspath{ {fig/} }

\begin{document}
\nolinenumbers
%%Temporary for christmas present
% Place the image in the top-right corner

\title{GlobalBuildingAtlas: An Open Global and Complete Dataset of Building Polygons, Heights and LoD1 3D Models}

%\title{~~~~~~L/ASMSS A Large Scale and Artisanal Mining Site Segmentation Dataset \ZC{Maybe some ML/RS keywords in the title?}}

% \Author[affil]{given_name}{surname}

\Author[1,2][xiaoxiang.zhu@tum.de]{Xiao Xiang}{Zhu}
\Author[1,2,\dag]{Sining}{Chen}
\Author[1,\dag]{Fahong}{Zhang}
\Author[1]{Yilei}{Shi}
\Author[1]{Yuanyuan}{Wang}

\affil[1]{Technical University of Munich, Munich, Germany}
\affil[2]{Munich Center for Machine Learning}
\affil[\dag]{These authors contributed equally to this work.}

\correspondence{Xiao Xiang Zhu (xiaoxiang.zhu@tum.de)}

%\affil[]{}

%% The [] brackets identify the author with the corresponding affiliation. 1, 2, 3, etc. should be inserted.

%% If an author is deceased, please mark the respective author name(s) with a dagger, e.g. "\Author[2,$\dag$]{Anton}{Smith}", and add a further "\affil[$\dag$]{deceased, 1 July 2019}".

%% If authors contributed equally, please mark the respective author names with an asterisk, e.g. "\Author[2,*]{Anton}{Smith}" and "\Author[3,*]{Bradley}{Miller}" and add a further affiliation: "\affil[*]{These authors contributed equally to this work.}".

\runningtitle{global3D}

\runningauthor{mk}

\received{}
\pubdiscuss{} %% only important for two-stage journals
\revised{}
\accepted{}
\published{}

%% These dates will be inserted by Copernicus Publications during the typesetting process.

\firstpage{1}

\maketitle

\begin{abstract}
We introduce GlobalBuildingAtlas, a publicly available dataset providing global and complete coverage of building polygons, heights and Level of Detail 1 (LoD1) 3D building models. This is the first open dataset to offer high quality, consistent, and complete building data in 2D and 3D form at the individual building level on a global scale. Towards this dataset, we developed machine learning-based pipelines to derive building polygons and heights (called GBA.Height) from global PlanetScope satellite data, respectively. Also a quality-based fusion strategy was employed to generate higher-quality polygons (called GBA.Polygon) based on existing open building polygons, including our own derived one. With more than 2.75 billion buildings worldwide, GBA.Polygon surpasses the most comprehensive database to date by more than 1 billion buildings. GBA.Height offers the most detailed and accurate global 3D building height maps to date, achieving a spatial resolution of 3×3 meters—30 times finer than previous global products (90 m), enabling a high-resolution and reliable analysis of building volumes at both local and global scales. Finally, we generated a global LoD1 building model (called GBA.LoD1) from the resulting GBA.Polygon and GBA.Height. GBA.LoD1 represents the first complete global LoD1 building models, including 2.68 billion building instances with predicted heights, i.e., with a height completeness of more than 97\%, achieving RMSEs ranging from 1.5 m to 8.9 m across different continents. With its height accuracy, comprehensive global coverage and rich spatial details, GlobalBuildingAltas offers novel insights on the status quo of global buildings, which unlocks unprecedented geospatial analysis possibilities, as showcased by a better illustration of where people live and a more comprehensive monitoring of the progress on the 11th Sustainable Development Goal of the United Nations. The code is publicly available at \url{https://github.com/zhu-xlab/GlobalBuildingAtlas}, and the GlobalBuildingAtlas dataset is available at \url{https://mediatum.ub.tum.de/1782307} (\cite{mediatum}).   
%    \\   
%\textbf{sources:} https://github.com/zhu-xlab \\
%\textbf{data-set:} https://github.com/zhu-xlab    
 
    %{please kindly add the github repository here. Here I suggest to use the lab github instead of the future lab one, as it is part of the DynamicEarthNet project. The source will be reviewed and added in a later stage}
  \end{abstract}

%\copyrightstatement{TEXT} %% This section is optional and can be used for copyright transfers.

\section{Introduction}  %% \introduction[modified heading if necessary]
Buildings anchor human life and define the form and function of urban environments. According to the United Nations (UN), over 50\% of the global population currently reside in cities and considering the ongoing urbanization, it is estimated that this ratio will reach nearly 70\% by 2050 (\cite{un2024sdg}). While urbanization enables more people to live in connected communities, it also introduces a range of challenges, including inequalities and urban poverty, inadequate transportation infrastructure, air pollution, and limited access to open public spaces. In response, the UN has established "sustainable cities and communities" as its 11th Sustainable Development Goal (SDG 11) in its 2030 Agenda. 

To monitor the progress toward this goal, one of the key indicators being used is the ratio of the land consumption rate to the population growth rate (Indicator 11.3.1, \cite{unsdg_indicator})---a ratio that measures how much land is being developed or consumed relative to the rate at which the population is growing.  This key indicator is computed based on the built-up area. Although this metric provides some insights into the spatial distribution of the built environment, its reliance on two-dimensional measurements can lead to biased or misleading interpretations. Cities are inherently three-dimensional (3D), and ignoring vertical space overlooks crucial information about how urban space is actually used. For example, a densely populated informal settlement may have the same built-up area per capita as a well-designed urban neighborhood with multi-story buildings, despite their vastly different spatial, social and infrastructural conditions. Therefore, a more complete understanding of the actual urban form requires information on building heights, which would enable obtaining a volumetric perspective of the built environment. Such 3D insights are essential for urban planning, infrastructure management and policy-making---especially in resource-limited contexts where the strategic allocation of funding and intervention is critical.

Despite its importance, the availability of comprehensive building height data remains unevenly distributed across the globe. While developed countries often leverage advanced technologies to monitor urban development, many regions undergoing rapid change---particularly in the Global South---lack the observational infrastructure and technical capacity to do so. Notably, these are often the areas that are most vulnerable and most in need of accurate, timely data. A promising alternative lies in Earth observations from space, which can offer scalable and repeatable insights independent of local ground-based or aerial-based resources.

However, existing global-scale building products utilizing data collected from space fall short in several key aspects. First, true global coverage and completeness remain limited, with most datasets covering only continental or national extents. Second, many existing approaches yield only coarse, aggregate-level building representations, which are insufficient for applications requiring more detailed, building-level information. Third, the reliance on ancillary data---such as socioeconomic variables, like population---hinders the scalability and timeliness of such product, especially in fast-changing urban contexts where rapid updates are critical. 

To address these limitations, we introduce GlobalBuildignAtlas (GBA), a global-scale dataset originating from a pipeline that relies exclusively on optical satellite imagery. GBA provides comprehensive building polygons, heights and level of detail 1 (LoD1) 3D building models. Our main contributions are
\begin{itemize}
  \item GBA.Polygon: The first complete set of global building polygons consisting of 2.75 billion buildings, closing the current gap of more than 40\% global buildings that were previously not accounted for. 
  \item GBA.Height: The most detailed and accurate global 3D building height map to date, achieving an unprecedented spatial resolution of 3×3 meters---30 times finer than previous global products (90 m), and enabling a high-resolution and reliable analysis of building volumes at both local and global scales, achieving root mean square errors (RMSEs) ranging from 38.0 $\text{m}^3/100~\text{m}^2$ to 580.0 $\text{m}^3/100~\text{m}^2$ across different continents.
  % \item global LOD1 building models with the highest (height?) completeness amount to 95 \% and height accuracy of xx meter. 
  \item GBA.LoD1: The first complete global LoD1 building model, including 2.68 billion building instances with predicted heights, which achieves RMSEs ranging from 1.5 m to 8.9 m across different continents.
\end{itemize}

% \begin{itemize}
%   \item Background and Significance: Explain the importance of urban research and the limitations of previous datasets; Highlight the gaps in data coverage and resolution that the Global3D dataset addresses.
%   \item Objective: creating a comprehensive global 3D building model; the potential of this dataset for advancing research in urban planning, environmental studies, disaster management, and more.
% \end{itemize}

% Key Contributions:

% \begin{itemize}
%   \item We present the most detailed and accurate global 3D building map to date, achieving an unprecedented spatial resolution of 3×3 meters—30 times finer than previous global products (90m)—with an average building height error below 4 meters. This enables a high-resolution and reliable analysis of building volume at both local and global scales.
%   \item The most complete global building polygons, closing the gap of xx\% global buildings (ours: 2.8 billion; 3D-GlobFP: 1.66 billion).
%   % \item global LOD1 building models with the highest (height?) completeness amount to 95 \% and height accuracy of xx meter. 
%   \item global LoD1 building models with RMSEs ranging from 3m (tbd, could be smaller) to 8m across 5 continents.

% \end{itemize}

%%%%%%%%%%%%%%%%%%%%%%%%%%%%%%%%%%%%%%%%%%%%%%%%%%%%%%
\section{Related Work}\label{background}
3D building information is crucial for many purposes, including urban planning. Such information captures structural attributes that extend beyond the ground footprint of the building, including the building height and volume. Table \ref{tab:related_works} summarizes existing mainstream products that primarily map 3D building information in two main aspects: raster-based and instance-level building representations.
\subsection{Raster-based Building Representation Products}
Raster-based building representations provide the statistics of buildings in regular grids. The products are delivered as rasters showing aggregated building heights or total building volumes in a grid format, typically spanning tens to hundreds of meters. The results are often estimated from diverse data sources using simple regression models. For example, World Settlement Footprint (WSF) 3D (\cite{Esch2020,Esch2022}) quantifies average building heights, total volumes and building fractions at 90 m resolution. Global Human Settlement Layers (GHSL) (\cite{Pesaresi2024}) model global building distributions, with GHS-BUILT-H (\cite{Pesaresi2023}) depicting building heights and GHS-BUILT-V (\cite{Pesaresi2023b}) depicting volumes at 250 m resolution. \cite{Ma2024} provided global building heights at a spatial resolution of 150 m. Generally, these products are derived either from low-resolution data sources of multiple modalities, such as Sentinel-1 Synthetic Aperture Radar (SAR) images and Sentinel-2 multi-spectral images, or from global digital elevation model (DEM) sources at coarse resolution. To achieve higher resolution mapping, recently Google (\cite{Sirko2023})
demonstrated the potential of optical satellite imagery for the rapid retrieval of building information by generating building presence and height data at a 4-meter resolution using only Sentinel-2 images. However, this dataset is not globally available yet and is primarily limited to regions in the Global South. 

In summary, raster-based products of this kind usually have a large coverage or even global coverage, but with the compromise of a lower quality and resolution, restricting their use in applications requiring high granularity at the instance level.

\subsection{Instance-level Building Representation Products}
Instance-level building representation maps delineate individual buildings and assign their corresponding 3D features. There already exist several 3D building model products at the instance level, albeit on a smaller scale (e.g., city or country). For instance, \cite{peters2022} created 3D-BAG, the first open 3D building dataset based on Light Detection and Ranging (LiDAR) data covering the Netherlands; while the \cite{Helsinki3D} released a high-quality 3D semantic city model derived from high-resolution aerial photographs. Similar models are available in other developed areas worldwide, e.g., Adelaide by the \cite{SA3DModel}, Philadelphia by \cite{Philadelphia3D}, and 260 cities in Japan by the \cite{PLATEAU3D}. Although they are derived from high-quality observations, and are capable of reconstructing detailed 3D building structures at Level of Detail 2 (LoD-2) or higher, such products are not widely available globally due to their high operational costs and significant computational demands.

In addition to the aforementioned observation-based techniques, Volunteered Geographic Information (VGI) offers an alternative approach. While VGI—based on human annotations—is often recognized for its positional accuracy, it typically suffers from limited completeness, particularly with respect to building height information. Therefore, current mainstream products are primarily derived from aerial or satellite remote sensing data using algorithmic approaches. Microsoft (\cite{MicrosoftBuildingFootprints}) maintains a dataset of 1.4 billion building footprints, of which approximately 20\% include estimated building heights. \cite{essd-16-5357-2024} provided 3D-GloBFP---a 3D building footprint dataset encompassing 1.66 billion buildings.
Despite achieving substantial coverage in terms of building footprints and height estimates, none of these datasets are derived from complete or globally consistent sources. Consequently, the total number of buildings represented falls significantly short of the United Nations' estimate of approximately 4 billion buildings worldwide (\cite{unhabitat2019}).
\subsection{Gaps in Existing Products}
While raster-based building representation products prioritize broader coverage, they suffer from low resolution and quality limitations. Instance-level representation products, on the other hand, offer detailed information on individual buildings but are constrained by limited coverage and incomplete datasets. Moreover, the production of high-level building representation products often involves multiple data sources, complicating the process of updating these datasets efficiently.

To address these gaps, we propose a novel pipeline to generate global LoD1 building models based solely on PlanetScope images. Leveraging this approach, we generated 3 m-resolution raster height maps alongside globally consistent building polygons. By fusing the generated building polygons with existing open-source building footprints based on quality assessments, we constructed a global building footprint dataset optimized for the highest possible quality. Finally, we assigned building heights to every individual building polygon using our generated height maps, which led to the first complete global LoD1 building models.  

\setlength\tabcolsep{0pt}
\begin{table}[h!]
\newcolumntype{C}{>{\centering \arraybackslash}m{1.2cm}}
\newcolumntype{D}{>{\centering \arraybackslash}m{2.2cm}}
\newcolumntype{E}{>{\centering \arraybackslash}m{2.7cm}}
\newcolumntype{F}{>{\centering \arraybackslash}m{3.5cm}}
\footnotesize
\centering
\caption{Comparison of existing large-scale building map and building height products. BF: Building footprint; BH: Building height; B: billion.}
\begin{tabular}{D D F D D C D E}
\toprule
Product & Data Structure & Data Sources & BF Coverage & BH Coverage & Resolution & Building Count / with Height & Year \\
\midrule
WSF 3D \cite{} & Raster & S1, S2, TanDEM-X  & Global & Global & 90m & / & vary from 2011-2019\\
GHS-BUILT-H \cite{}  & Raster & 5 Global DEM sources & Global & Global & 250m & / & vary from 2000-2011\\
GBH 2020 \cite{} & Raster & LS-8, S1, S2 & Global & Global & 150m & / & vary from 2019-2021\\
Google 2.5D \cite{} & Raster & S2 & Global South & Global South & 4m & / & 2016 - 2023 Yearly \\
GBA.Height (ours) & Raster & PlanetScope & Global & Global & 3m & / & 2019 \\
\addlinespace[0.5ex]
\midrule
\midrule
\addlinespace[0.5ex]
Microsoft & Vector & Multi-source VHR images & Global excl. China & NA, EU, OC & / & 1.4 B / 0.28 B & vary from 2014-2021 \\
OSM & Vector & Human Annotation & Global in part &  Global in part & / & 0.45 B/ 0.02 B & since 2008 \\
% Open Buildings & Vector & multi-source VHR images & Global South & / & / & 1.8 B & until 2021  \\
% CLSM & Vector & Google Earth images & East Asia & / & / & 0.28 B & vary from 2020-2022 \\
3D-GloBFP & Vector & 10 different sources & Global & Global & / & 1.7 B / 1.7 B & vary from 2014-2021 \\
GBA.LoD1 (ours) & Vector & 4 BF sources, PlanetScope & Global & Global & / & 2.75 B / 2.68 B & 2019 \\

\bottomrule
\end{tabular}
\label{tab:related_works}
\end{table}

%%%%%%%%%%%%%%%%%%%%%%%%%%%%%%%%%%%%%%%%%%%%%%%%%%%%%%
\section{Data Sources}\label{source}
This section introduces all the data sources that were used to develop the GlobalBuildingAtlas dataset.
\begin{figure}[ht]
    \centering
    \includegraphics[width=1.0\textwidth]{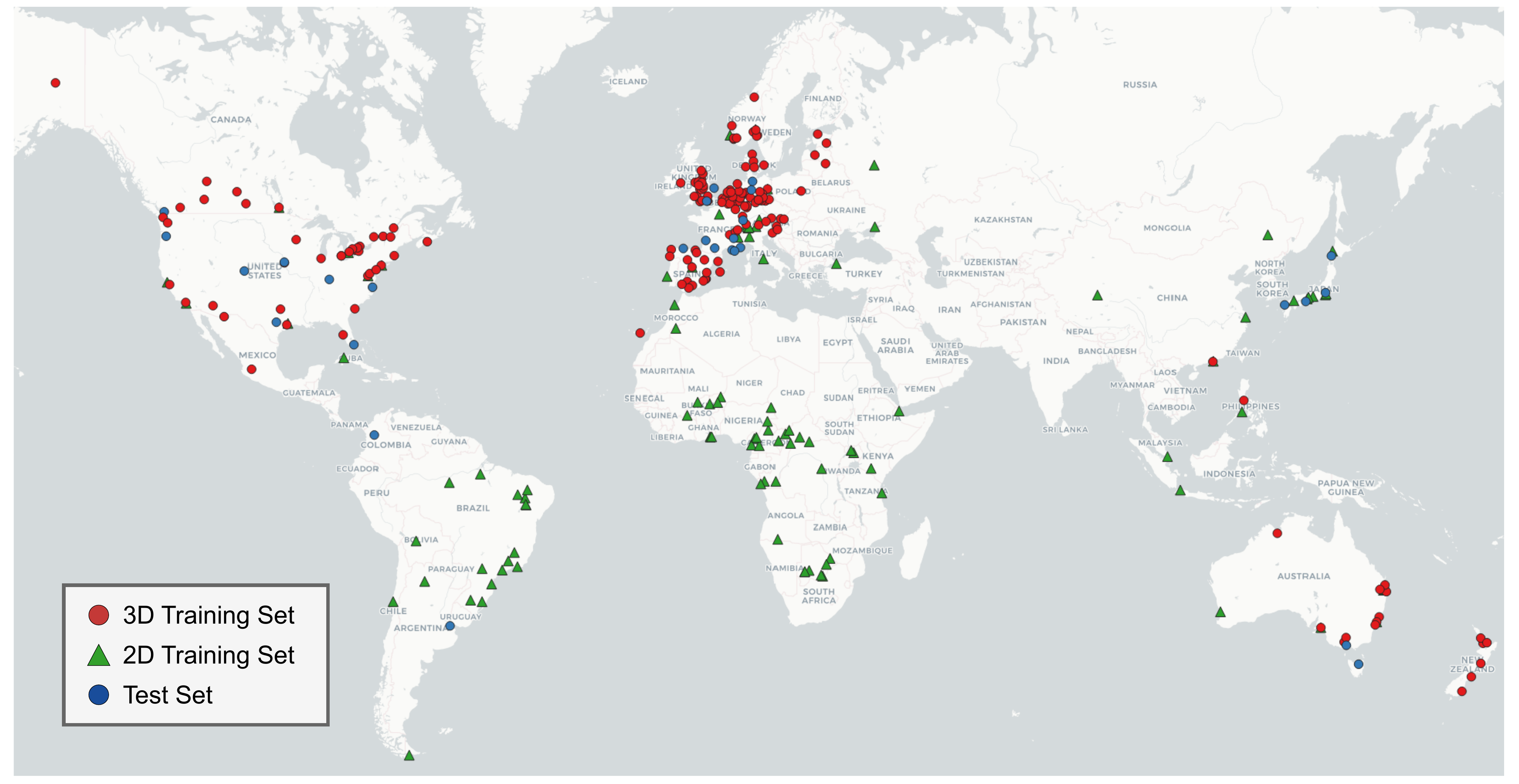}
    \caption{Distribution of city-scale regions of interest where training and test data were collected. The 3D and 2D training sets were used to train the building height estimation and building polygon generation pipelines, respectively.}
    \label{fig:data_distribution}
\end{figure}

\subsection{Satellite Optical Images} % Planet product charateristics
We collected global-scale PlanetScope Surface Reflectance (PSR) imagery to support polygonal building mapping and height estimation. PSR data are orthorectified, multi-spectral satellite images provided by Planet's satellite constellation. These images undergo atmospheric correction, ensuring that the reflectance values accurately represent the Earth's surface properties. The PSR data we utilized include four spectral bands—three in the visible RGB range and one near-infrared (NIR) band, with a spatial resolution of approximately 3 meters. Additionally, the PlanetScope constellation offers a high temporal revisit frequency, capturing imagery of the same location up to daily, which is beneficial for monitoring dynamic urban environments.

\subsection{Building Footprints}
We utilized building footprint datasets from various sources (see Table \ref{tab:existing_bfs}), including OSM (\cite{OpenStreetMap}), Google Open buildings (\cite{GoogleOpenBuildings}), Microsoft Building Footprints (\cite{MicrosoftBuildingFootprints}), and CLSM (\cite{Shi2024GlobalBuildingFootprints}). Since none of the above-mentioned footprints is complete, we also generated our own global building polygons from an updated version of GlobalBuildingMap (\cite{zhu2024globalbuildingmapunveilingmystery}), with a novel building regularization and polygonization pipeline applied. The building mapping and regularization networks were trained on PSR imagery using manually filtered OpenStreetMap (OSM) annotations with high completeness. The data distribution is illustrated in Fig.~\ref{fig:data_distribution}.
To produce the final complete global building footprint dataset---GBA.Polygon---we integrated all available sources, including our own generated polygons, using a quality-based fusion approach.

\begin{table}[h!]
\newcolumntype{C}{>{\centering \arraybackslash}m{1.2cm}}
\newcolumntype{D}{>{\centering \arraybackslash}m{2.2cm}}
\newcolumntype{E}{>{\centering \arraybackslash}m{2.7cm}}
\newcolumntype{F}{>{\centering \arraybackslash}m{3.5cm}}
\footnotesize
\centering
\caption{Comparison of existing large-scale building footpring products. B: billion.}
\begin{tabular}{E D E F}
\toprule
Product & Coverage  & Number of Buildings & Year \\
\midrule
\addlinespace[0.5ex]
Microsoft & Global excl. China & 1.4 B & vary from 2014-2021 \\
OSM  & Global in part & 0.45 B & since 2008 \\
Open Buildings & Global South & 1.8 B & until 2021  \\
CLSM & East Asia & 0.28 B & vary from 2020-2022 \\
3D-GloBFP & Global & 1.7 B & vary from 2014-2021 \\
GBA.Polygon (ours) & Global & 2.75 B & vary from 2014-2021 \\

\bottomrule
\end{tabular}
\label{tab:existing_bfs}
\end{table}

\begin{comment}
As an auxilary source, it's maybe better to not mention it here?
\subsection{Land Cover Maps}
The above mentioned building footprint products 

Wolrd Cover 2020 from \cite{zanaga2021worldcover,zanaga2022worldcover} is used to filter the false alarms generated by the proposed building polygon generation framework. % More details by Fahong?
\end{comment}

\subsection{LiDAR}
As the most precise 3D measurement available, we used LiDAR data covering a wide range of geolocations (see Fig. \ref{fig:data_distribution}) in developed countries. These data are publicly released by governments for open use. Due to the high operational cost of LiDAR observations, no such data are available in Africa. We processed the LiDAR data into normalized digital surface models (nDSMs), which represent the heights of ground objects. These nDSMs served as reference data for training our neural networks.

%%%%%%%%%%%%%%%%%%%%%%%%%%%%%%%%%%%%%%%%%%%%%%%%%%%%%%
\section{Methodology}\label{Methodology}
\subsection{Overall Workflow}
\begin{figure}[ht]
    \centering
    \includegraphics[width=1.0\textwidth]{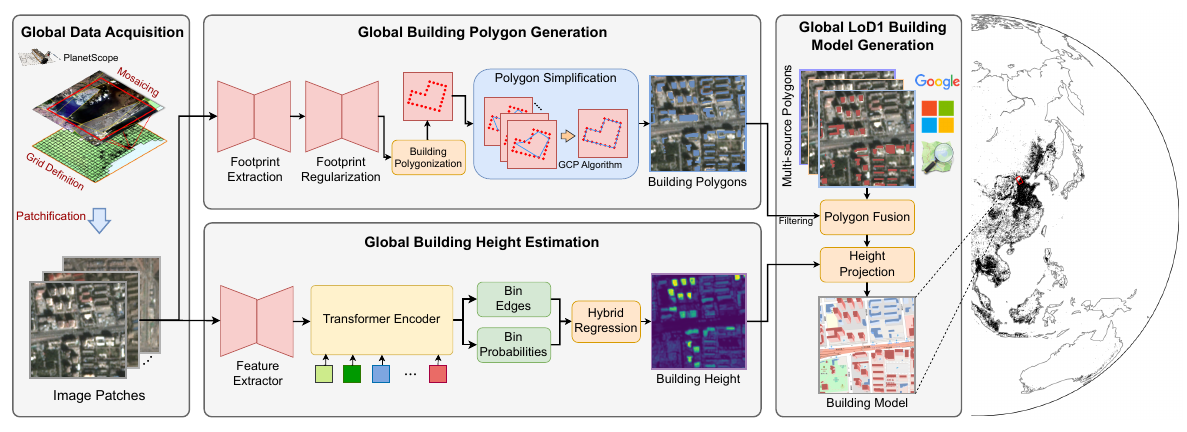}
    \caption{Workflow of the proposed pipeline.}
    \label{fig:methdology}
\end{figure}

The overall workflow for generating the GlobalBuildingAtlas dataset is illustrated in Fig. \ref{fig:methdology}.
The workflow can be divided into $4$ sections,
including global data acquisition, global building polygon generation, global building height estimation and post processing, i.e., global LoD1 model generation.

\subsection{Global Data Acquisition}
\label{sec:data_acquire}
This section details the acquisition, organization and preprocessing of PSR data to prepare them for the subsequent deep learning models and LoD1 model generation.
We divided the Earth's surface into grid cells of $0.2^\circ \times 0.2^\circ$ . Grids overlapping with built-up areas, as defined by the Global Urban Footprint (GUF) dataset (\cite{esch2010delineation,esch2011characterization,esch2012tandem,esch2013urban}), were selected as our areas of interest. For these regions, we acquired satellite imagery from 2019 that intersects with the selected grids, aiming to maximize coverage.
In total, approximately 800,000 PlanetScope scenes were downloaded, each with a spatial resolution of 3 meters and covering an area of about 287.5 km$^2$.

To ensure data quality, we applied a cloud filter that retained only scenes with less than 10\% cloud cover. In areas where suitable cloud-free imagery from 2019 was unavailable, the dataset was supplemented with images from 2018.
For each $0.2^\circ \times 0.2^\circ$ grid cell, PSR imagery was mosaicked using a scene prioritization strategy guided by the corresponding Unusable Data Masks (UDMs), ensuring the selection of the clearest and most complete pixels across overlapping scenes.

\subsection{Global Building Polygon Generation}
\label{sec:polygon_generation}

This section outlines the global-scale building polygon generation process, as illustrated in the upper middle part of Fig. \ref{fig:methdology}. We begin by describing the curation of the training data, followed by a detailed description of the building footprint extraction model, the regularization model, and the polygonization method used to generate the building polygons. Finally, we present the filtering process employed to remove false-positive polygons.

\subsubsection{Data Curation}
% Elaborate the gap of existing products.
To construct a dataset for training the building polygon generation model, we sampled \gls*{PSR} image patches from 107 predefined regions of interest (RoIs) distributed globally. These RoIs included 92 densely built urban areas and 15 non-urban regions, such as ocean, forest, desert, and mountain environments, which served as negative samples.
% The geographic distribution of these RoIs is shown in Fig. \ref{fig:data_distribution}. Move to appendix

PSR imagery overlapping with the RoIs was divided into patches of size $256 \times 256$ to ensure compatibility with the deep learning models, resulting in a total of 142,722 training samples.
For each sampled \gls*{PSR} image patch, we obtained building polygons from \gls*{OSM} for regions outside of China. For Chinese cities, we used annotations provided by the dataset published by \cite{cao2021deep}.
All the acquired building polygon annotations were rasterized to the spatial resolution of $3$ meter to match the resolution of the \gls*{PSR} images.

\subsubsection{Building Map Extraction}
\label{sec:bf_extraction}
A building map extraction network was designed to map the input \gls*{PSR} images to binary building masks. This followed an encoder–decoder architecture based on UPerNet (\cite{xiao2018unified}), with ConvNeXt-Tiny (\cite{liu2022convnet}) serving as the backbone. To facilitate deep supervision during training, an auxiliary decoder based on a Fully Convolutional Network (FCN) (\cite{long2015fully}) was added. Both the main and auxiliary decoder heads were supervised using the cross-entropy loss.
Before being fed into the network, image patches were upscaled by a factor of 4 and cropped to a size of $512 \times 512$. The network was trained for 160,000 iterations with a batch size of 8, using the AdamW optimizer (\cite{loshchilov2017decoupled}). The global inference was built-up on the pipeline for generating GlobalBuildingMap, as described in \cite{zhu2024globalbuildingmapunveilingmystery}.

\subsubsection{Building Map Regularization}
The building map extraction network described in Sect. \ref{sec:bf_extraction} generates binary building masks from input satellite images by applying a threshold of 0.5 to the predicted probability maps. However, the resulting masks are often noisy, with adjacent buildings merging due to the limited resolution of the \gls*{PSR} images, which are typically insufficient to capture fine building details. To address this, we trained a building map regularization network to refine the generated masks to improve their accuracy and delineation.

During training, we generated two binary masks, \( \hat{\mathbf{M}} \) and \( \mathbf{M} \), using building polygon annotations, treating them as the network’s input and output, respectively. \( \hat{\mathbf{M}} \) was created by rasterizing the building polygon annotations to match the resolution of the corresponding images, with random noise added to the vertices and edges of the polygons. In contrast, \( \mathbf{M} \) was generated similarly, but without any added noise. By learning to predict \( \mathbf{M} \) from \( \hat{\mathbf{M}} \), the network was able to denoise the generated binary masks, producing regularized and more accurate building mask representations.

Since both networks followed an encoder-decoder architecture, the architecture and training details of the building map regularization network were identical to those of the building map extraction network described in Sect. \ref{sec:bf_extraction}.

\subsubsection{Building Polygonization and Simplification}
After the regularized binary building maps were generated, the subsequent building polygonization and simplification algorithms mapped them to polygonal representations. Specifically, the binary masks were first converted into dense vector representations using a contour tracing-like algorithm provided by the GDAL library (\cite{gdal}). Then, a polygon simplification algorithm, as described in \cite{GCP}, was applied to simplify the polygonal representation.

\subsubsection{False Positives Filtering}
Since the available training data do not sufficiently cover the diverse terrains and land cover types worldwide, and the quality of satellite data can vary by region, the generated building polygons may include false positives, particularly in areas with cloud cover, snow, water bodies, forests, and similar features. To filter out these false positives, we implemented a filtering strategy based on the global land cover map product World Cover (\cite{zanaga2021worldcover}). We began by dilating the built-up mask of the World Cover map with a 250-meter window size, and then filter out any generated polygons that fell outside the masked area.

\subsection{Global Building Height Estimation}
\label{sec:height_estimation}

The proposed pipeline acquires 3D building information exclusively from optical satellite imagery, ensuring broad accessibility and ease of data acquisition. We utilized deep learning methods that are particularly well-suited for this task, as they can effectively leverage large-scale data while offering flexibility and scalability—key advantages for global geospatial applications. This section details the height information retrieval process, illustrated in the lower middle part of Fig. \ref{fig:methdology}. We first describe the curation of the training data and then elaborate on the monocular height estimation process.
\subsubsection{Data Curation}
The dataset used for training the monocular height estimation model consisted of samples from 168 city-scale RoIs worldwide (see Fig. \ref{fig:data_distribution}).
% as shown in Fig. \ref{fig:data_distribution}. Move to appendix
These regions were primarily located in North America, Europe and Oceania, where high-quality 3D observations are more readily available. The 3D data orginated from LiDAR point clouds released by governments, which were processed into height maps in raster format---specifically, normalized digital surface models (nDSM) that match the resolution of PSR imagery. The nDSMs were cropped into patches of size 256 $\times$ 256, yielding a total of 231,656 samples.
\subsubsection{Monocular Height Estimation}
We trained an HTC-DC Net (\cite{htcdc}) to predict heights from single images, following a classification-regression paradigm. The model consisted of an EfficientNet-B5 (\cite{tan2019efficientnet}) backbone for feature extraction, a classification module, and a hybrid regression process. The classification module employed a vision transformer encoder (\cite{dosovitskiy2020image}) to capture relationships between local and global features, dynamically determining bin edges and their corresponding probabilities. The classification output is treated as a distribution, which was refined through the hybrid regression process to obtain the final height prediction. The network was trained for 150 epochs with a batch size of 8, using the AdamW optimizer (\cite{loshchilov2017decoupled}).

\subsubsection{Uncertainty Quantification}
To assess the uncertainty of the predicted height values, we employed the test-time augmentation (TTA) technique during the inference process. Inference was conducted on mosaicked PSR imagery on a $0.2^\circ \times 0.2^\circ$ scale using a sliding window approach. The window slides with a stride of 128 pixels, allowing up to 4 predictions per pixel. The variance across these 4 predictions serves as the uncertainty measure for height prediction.

\subsection{Global LoD1 Building Model Generation}
The generation of the global LOD1 building model relies on the availability of high-quality global building polygons and the corresponding building heights. 
Given the limited availability of accurate and detailed global building height datasets, GBA.Height represents the most suitable option for this purpose. However, to leverage existing large-scale building footprint datasets---such as the vector sources listed in Table \ref{tab:existing_bfs}, which were derived from higher-quality input data---we introduce a quality-guided fusion strategy. This approach, detailed in Sect. \ref{sec:quality_guided}, enables the integration of multiple building polygon sources. The final global LoD1 building model was then constructed based on this fused footprints and our generated GBA.Height, as described in Sect. \ref{sec:LOD1}.

\subsubsection{Quality-guided Building Polygon Fusion}
\label{sec:quality_guided}

As shown in Table \ref{tab:existing_bfs}, different instance-level building footprint products exhibit varying degrees of completeness across different regions. Even in areas where multiple building footprint products are available, their quality differs significantly. To leverage the strengths of all available datasets, we performed a quality-guided polygon fusion process to produce a final global building footprint dataset---GBA.Polygon.

The fusion processing was carried out within each administration boundary as defined by the GADM dataset (\cite{gadm2025}). The process begins with selecting a base layer, i.e., the primary source, which should represent the highest-quality building footprint at a given location. \cite{zhu2024globalbuildingmapunveilingmystery} compared various building footprint datasets across different continents and concluded that OpenStreetMap (OSM) is the most suitable base layer for all continents except South America and Africa. In these two regions, Google's Open Building dataset outperforms others due to its well-delineated and superior preservation of details.

However, while these datasets may provide the best quality at a large scale, concerns about their completeness remain. To address this limitation, we introduced a secondary source to complement the primary dataset with region-specific coverage.

To address this, we evaluated other sources against the base layer to identify the most suitable secondary source, with the ranking determined using a combined metric of recall and area gain. Recall measures the building areas in the primary source that are covered by a given secondary source, while area gain quantifies the additional building area introduced by the secondary source. The source with the highest combined metric was selected as the secondary source, ensuring that it aligned well with the primary source while maximizing the inclusion of new building footprints.

Finally, the selected primary and secondary sources were merged at the instance level. All buildings from the primary source were retained, and any additional buildings from the secondary source were incorporated into the final dataset. This approach ensures that the final polygon set integrates the strengths of the two most reliable data sources, resulting in a more complete and accurate global building footprint product.
\subsubsection{LoD1 Building Model Generation}
\label{sec:LOD1}
To generate the final LoD1 building models, the predicted height maps were integrated with the fused building footprints to assign a height value to each building instance. This height value was determined by selecting the maximum height within each building's footprint. The variance at the location where the maximum value was taken served as the uncertainty measure for the corresponding building instance.

%%%%%%%%%%%%%%%%%%%%%%%%%%%%%%%%%%%%%%%%%%%%%%%%%%%%%%%%%%%%%
% Move to supplementary?
%\section{Results}\label{datasets}
%\subsection{Dataset Overview}

\begin{figure}[ht]
    \centering
    \includegraphics[width=1.0\textwidth]{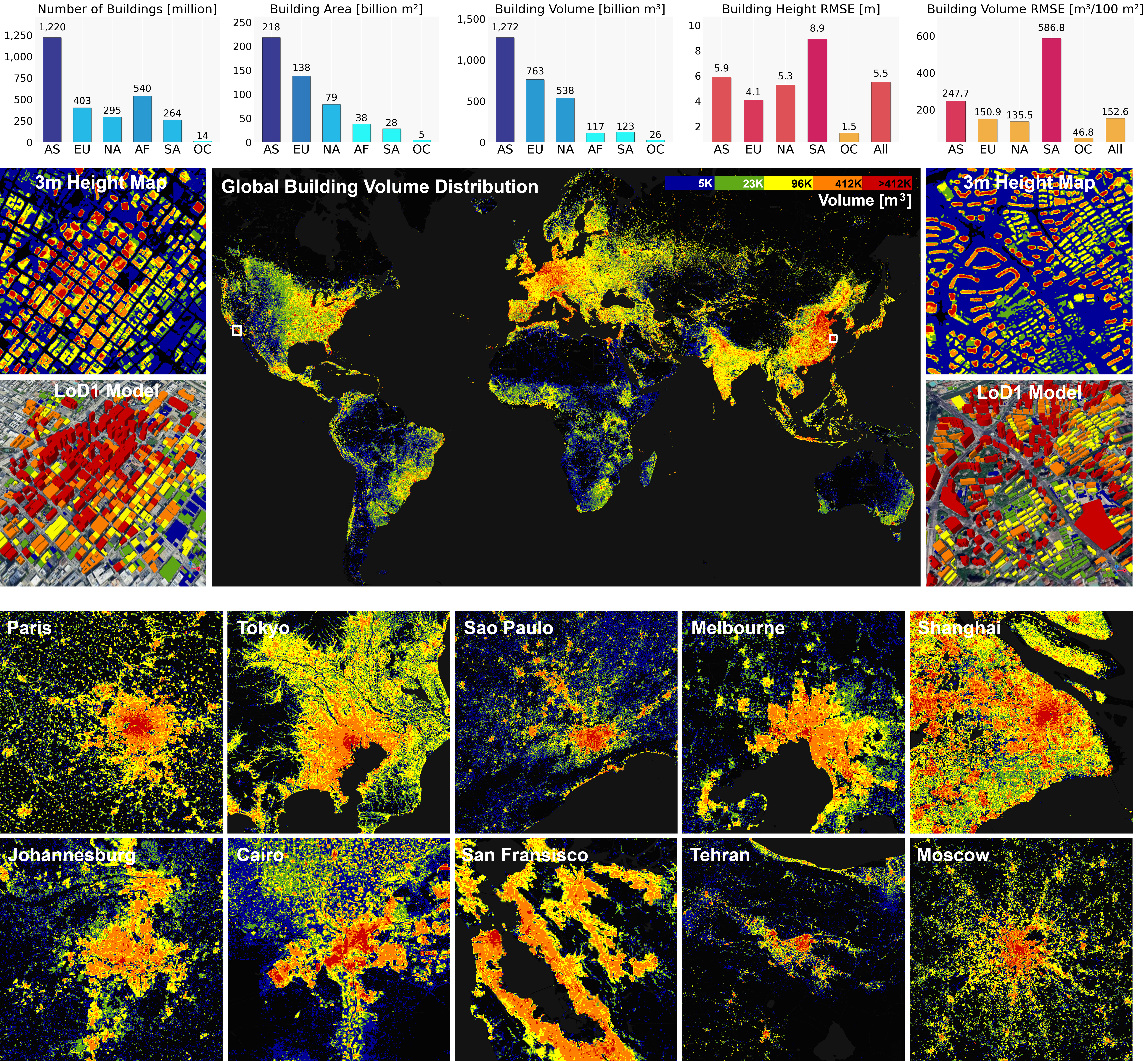}
    \caption{Overview of our resulting GlobalBuildingAtlas dataset, consisting of global building polygons (GBA.Polygon), building height map (GBA.Height) and LoD1 models (GBA.LoD1). The top section presents the dataset statistics by continent, including the number of buildings, total building area and volume, as well as height RMSE and volume RMSE across the test cities.
The middle section displays the global distribution of building volume, computed over $480~\text{m} \times 480~\text{m}$ grid cells.
The bottom section offers a closer look at the building volume distribution in 10 representative cities, using the same colormap as the global map for consistency.}
    \label{fig:overview}
\end{figure}

\begin{comment}
\iffalse 
\clearpage
\begin{figure}[ht]
    \ContinuedFloat
    \centering

    \includegraphics[width=1.0\textwidth]{figures/figure2_YW.pdf}
    
    \caption[]{ALTERNATIVE COLOR. Overview of our resulting GlobalBuildingAtlas dataset, consisting of global building polygons (GBA.Polygon), building height map (GBA.Height) and LoD1 models (GBA.LoD1). The top section presents the dataset statistics by continent, including the number of buildings, total building area and volume, as well as height RMSE and volume RMSE across the test cities.
The middle section displays the global distribution of building volume, computed over $480~\text{m} \times 480~\text{m}$ grid cells.
The bottom section offers a closer look at the building volume distribution in 10 representative cities, using the same colormap as the global map for consistency.}
    \label{fig:overview}
\end{figure}

\fi
\end{comment}

\section{Results}\label{datasets}
\subsection{Dataset Overview}
Our resulting GlobalBuildingAtlas dataset consists of global building polygons (GBA.Polygon), building height maps (GBA.Height) and LoD1 models (GBA.LoD1). Figure \ref{fig:overview} gives an overview on the GlobalBuildingAtlas dataset. It illustrates the global distribution of building volumes (in the middle) as well as statistics on the building counts, areas and volumes and associated errors by continent (top), and showcases the resulting building height map and LoD1 model from selected areas in North America (middle-left) and Asia (middle-right), respectively. The bottom subfigures offer a closer look at the
building volume distribution in 10 representative cities across the globe, using the same colormap as the global map for consistency. The total building counts across the globe is 2.75 billion, corresponding to a total building area of 506.64 billion square meters ($\text{m}^2$) and a total building volume of 2.85 trillion cubic meters ($\text{m}^3$). This suggests that the actual global building count is lower than the UN's estimation of 4 billion. As anticipated, building volumes are concentrated in metropolitan regions, with notable clusters in East Asia, Europe, and North America. 

The global distribution of building-related metrics by continent reveals significant geographic variations in terms of quantity and scale. 

Asia leads in nearly all metrics, with an estimated 1.22 billion buildings, the highest among all the continents. This reflects both the continent's large population and the extensive urban development across countries such as China, India, and those in Southeast Asia. Africa follows with approximately 540 million buildings, surpassing Europe (403 million) and North America (295 million) in terms of building count. South America contributes around 264 million buildings, while Oceania has the lowest count at 14 million buildings. 

Despite Africa’s relatively high number of buildings, its total building area is significantly lower, amounting to only 38 billion square meters ($\text{m}^2$), in contrast to Asia's 218 billion~$\text{m}^2$, Europe's 138 billion~$\text{m}^2$, and North America's 79 billion~$\text{m}^2$. South America and Oceania contribute 28 billion~$\text{m}^2$ and 5 billion~$\text{m}^2$, respectively. This disparity indicates that African buildings tend to be smaller in size, possibly reflecting the dominance of low-rise or informal structures---an expected pattern in less developed or peri-urban areas.

A similar pattern can be observed with the building volume statistics. Asia again ranks highest, with a cumulative building volume of approximately 1.272 trillion~$\text{m}^3$, followed by Europe (763 billion~$\text{m}^3$) and North America (538 billion~$\text{m}^3$). Africa, despite its large number of structures, contributes only 117 billion~$\text{m}^3$, further underscoring the prevalence of small-scale or single-story buildings. South America and Oceania contribute 123 billion~$\text{m}^3$ and 26 billion~$\text{m}^3$, respectively.

The building height and volume estimation accuracy of the GlobalBuildingAtlas dataset across continents is mainly influenced by several factors: (1) variations in building morphology, (2) the representation of regional characteristics in the training data, and (3) the comprehensiveness and quality of the validation data. The global average height RMSE of GBA.LoD1 stands at 5.5 meters (m). The lowest error is observed in Oceania (1.5 m), followed by Europe (4.1 m), North America (5.3 m), and Asia (5.9 m). South America, however, exhibits a significantly higher height RMSE of 8.9 m, suggesting a greater uncertainty or variation in building morphology, possibly due to a mixture of informal and vertical developments in urban regions. The global average volume RMSE of GBA.LoD1 is approximately 152.6~$\text{m}^3$ per 100~$\text{m}^2$ ($\text{m}^3/100~\text{m}^2$). South America again shows the highest error at 586.8~$\text{m}^3/100~\text{m}^2$, which may be attributed to compounding uncertainties in both the building footprint and height estimation. Asia follows with 247.7 $\text{m}^3/100~\text{m}^2$, then Europe (150.9~$\text{m}^3/100~\text{m}^2$), North America (135.5~$\text{m}^3/100~\text{m}^2$), and Oceania with the lowest error at 46.8~$\text{m}^3/100~\text{m}^2$. Furthermore, GBA.Height enhances building volume estimation in South America and Asia, achieving RMSE values of 580.0~$\text{m}^3/100~\text{m}^2$ and 38.0~$\text{m}^3/100~\text{m}^2$, respectively.

In summary, the statistics reflect significant continental differences in the scale of the built environment and estimation accuracy. While Asia dominates in terms of the building count, area, and volume, Africa’s high building count but low aggregate volume and area underscores the heavy presence of small-scale and potentially informal constructions. Conversely, Europe and Oceania demonstrate relatively lower estimation errors, suggesting a higher consistency in terms of the building structures and their better representation in the training data. South America, despite its moderate building count, emerges as a region with notably high estimation errors, indicating potential challenges in urban morphology or limitations in the training data. 

These findings provide valuable insights for global-scale urban modeling, infrastructure planning, and geospatial analytics.

%As anticipated, building volumes are concentrated in metropolitan regions, with notable clusters in East Asia, Europe, and North America. In terms of continents, as expected Asia has the highest number, area, and volume of buildings, while Oceania represents the smallest share of global buildings. Although Africa containd more buildings than Europe and North America, it lags behind in terms of building area and volume, suggesting a higher prevalence of small-scale structures and informal settlements -- an expected pattern in less developed regions. 

%%%%%%%%%%%%%%%%%%%%%%%%%%%%%%%%%%%%%%%%%%%%%%%%%%%%%%%%%%%%%

\subsection{Comparison with Existing Products}\label{comparison}
The released GlobalBuildingAtlas dataset provides comprehensive information for both raster-based and instance-level building representation. For comparison, we compared it against three raster-based products, namely GHS-BUILT-H (\cite{Pesaresi2023b}), WSF-3D (\cite{Esch2022}), and Google 2.5D (\cite{Sirko2023}), as well as two instance-level products, namely Microsoft (\cite{MicrosoftBuildingFootprints}) and 3D-GloBFP (\cite{essd-16-5357-2024}).

\subsubsection{Validation Dataset}
We collected a test dataset (see Fig. \ref{fig:data_distribution}) comprising LoD1 building data from government sources across 28 cities in Asia, Europe, North America, South America, and Oceania to evaluate the performance of the various building height products. Note, owing to the lack of reference data in Africa, it was not possible to conduct a quantitative evaluation in that region.

\subsubsection{Metrics}

The evaluation covered assessments of both 2D building footprint and 3D building information, 
The quality of 2D building footprints was quantified using the following metrics.

\begin{itemize}
    \item Intersection over union (IoU).
    IoU was used to evaluate both the raster-based and instance-level building products.
    For evaluating raster-based building map, we rasterized the reference polygons to 3 m resolution, resizing the building map to the same resolution and then calculated the IoU of the building area.
    For the instance-level building products, the IoU was calculated at the polygon level.
    
    \item Average precision of building polygons at an IoU threshold of 0.5 ($\text{AP}^{poly}_{50}$).
    To evaluate the quality of building polygons with vectorized building data, we adopted the standard average precision (AP) metric, which is widely used in object detection and instance segmentation (\cite{everingham2010pascal}). Specifically, we computed the AP for building polygons using an IoU threshold of 0.5, denoted as $\text{AP}^{poly}_{50}$, following the classic evaluation protocol.

    \item Average recall of building polygons at an IoU threshold of 0.5 ($\text{AR}^{poly}_{50}$). In addition to evaluating the average precision ($\text{AP}^{poly}_{50}$), we also computed the average recall (AR) to assess how comprehensive the dataset’s ability to comprehensively capture building instances is. Specifically, $\text{AR}^{poly}_{50}$ was determined to measure the recall performance of detected building polygons at an IoU threshold of 0.5, following the standard evaluation protocol.

    \item N-ratio of the detected building count. To evaluate the accuracy of the different instance-level building products, we employed the N-ratio metric, defined as the ratio between the number of detected buildings and the corresponding ground truth count. This metric provides a straightforward measure of how well the predicted building counts align with reference data. An N-ratio closer to 1 indicates better performance.

\end{itemize}

The assessment of the 3D building information included estimated building volume and height accuracy, corresponding to the instance-level and raster-based building representations, respectively. 
\begin{itemize}
    \item Root mean square error ($\text{RMSE}_{BV}$) and mean absolute error ($\text{MAE}_{BV}$) of the estimated building volume. The ground truth building models and the instance-level building datasets were rasterized. The RMSE and MAE were computed on a per-pixel basis with a 1m $\times$ 1m grid resolution.
    \item RMSE ($\text{RMSE}_{BH}$) and MAE ($\text{MAE}_{BH}$) of the estimated building heights. Building instances in the ground truth and in the various datasets were matched according to the maximum overlapping area. The RMSE and MAE of the building height were then computed based on these matched pairs.
    \item Building height completeness (Comp.). This metric represents the proportion of ground truth buildings (in numbers) for which valid predictions are available. A prediction is considered valid if it exceeds a minimum building height threshold, which was set to 1 meter in this study.
\end{itemize}

\subsubsection{Quantitative Results}
\begin{figure}[ht]
    \centering
    %\includegraphics[width=0.7\textwidth]{figures/radar_SC.pdf}
%\includegraphics[width=0.19\textwidth]{figures/Asia_rada_crop.pdf}
%\includegraphics[width=0.19\textwidth]{figures/Europe_radar_crop.pdf}
%\includegraphics[width=0.19\textwidth]{figures/North America_radar_crop.pdf}
%\includegraphics[width=0.19\textwidth]{figures/Oceania_radar_crop.pdf}
%\includegraphics[width=0.19\textwidth]{figures/South America_radar_crop.pdf}

%\includegraphics[width=0.4\textwidth]{figures/radar_legend.pdf}

% or not 5 continents in a row, in two rows

\includegraphics[width=0.3\textwidth]{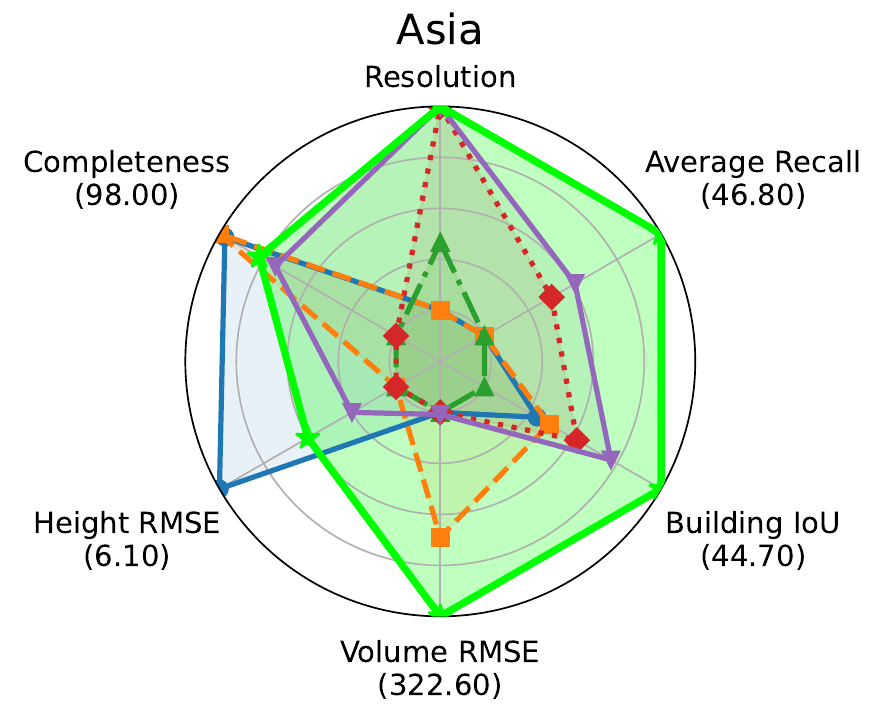}
\includegraphics[width=0.3\textwidth]{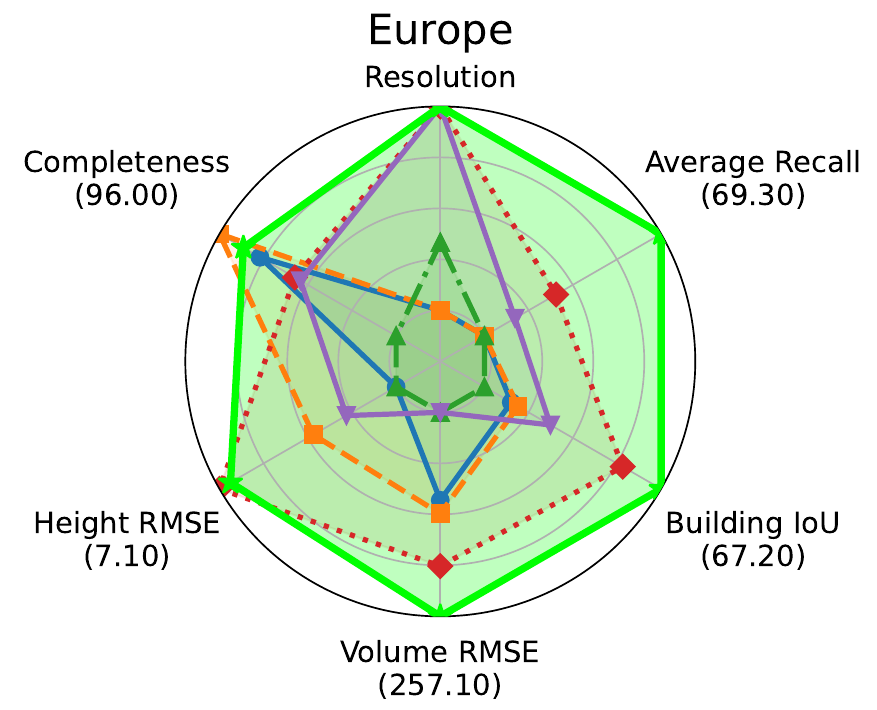}
\includegraphics[width=0.3\textwidth]{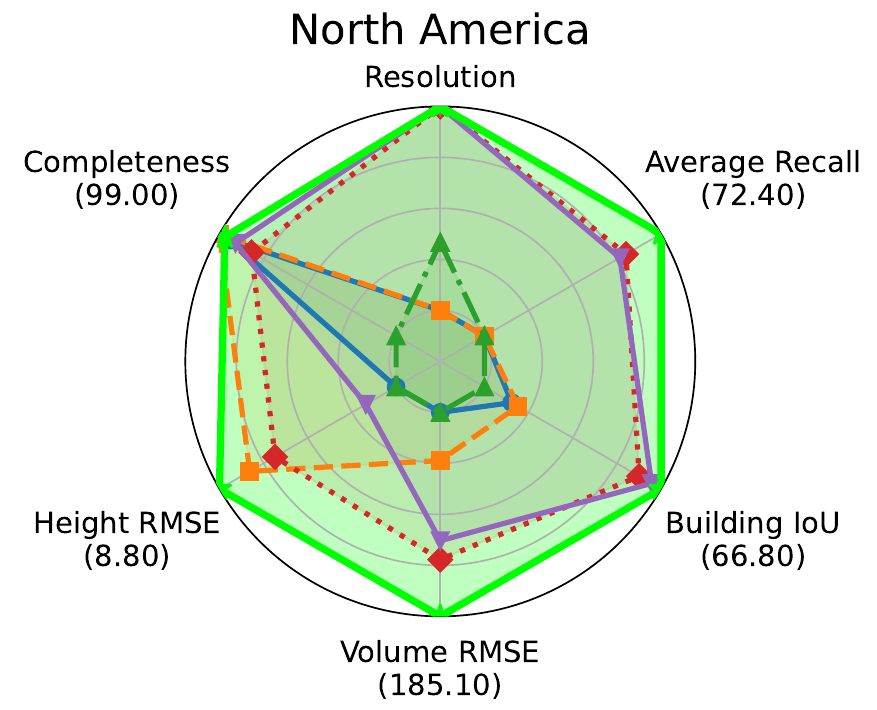}

\includegraphics[width=0.3\textwidth]{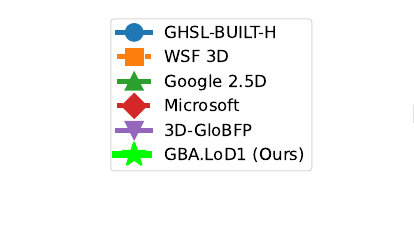}
\includegraphics[width=0.3\textwidth]{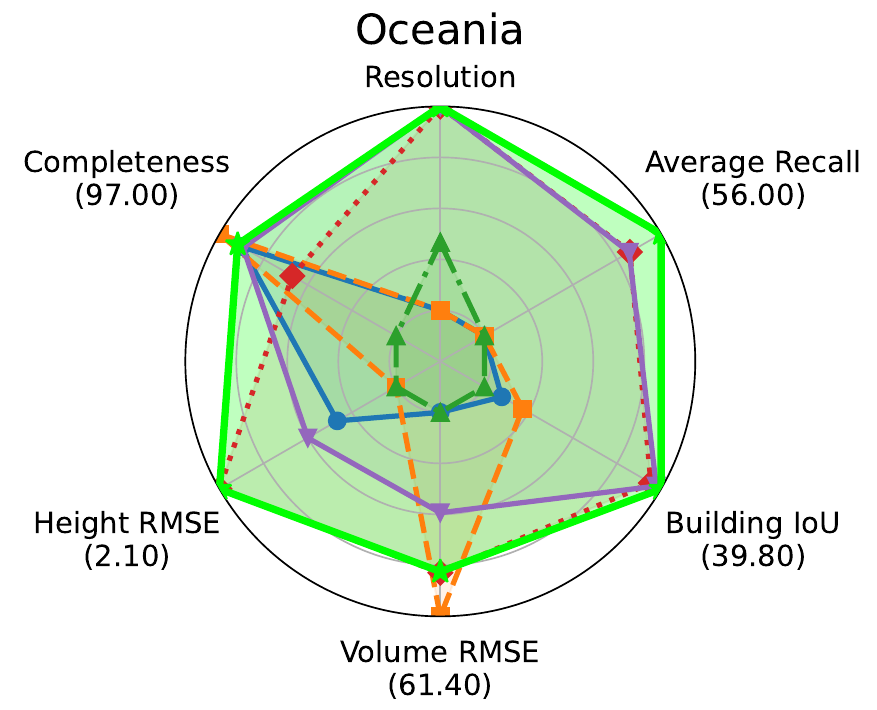}
\includegraphics[width=0.3\textwidth]{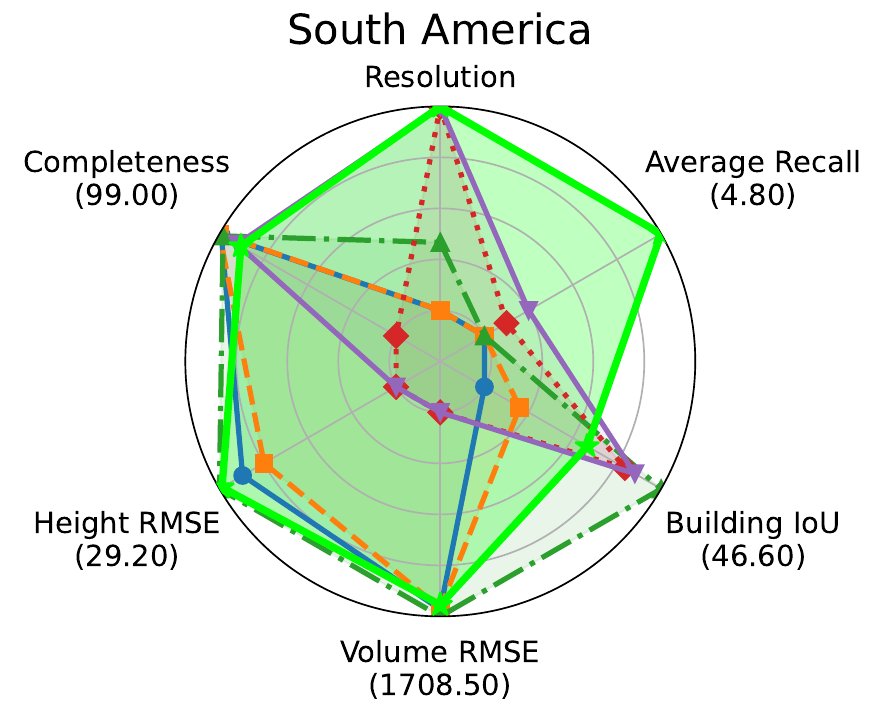}

    \caption{Overview of the continental comparison results across large-scale building height products. In the resolution dimension, the products were first scored based on whether they provide vectorized building footprints. Raster-based products were then further scored according to their spatial resolution.}
    \label{fig:radar_chart}
\end{figure}

We present the quantitative comparison results of various building height products in Table \ref{tab:comparison_results}. To provide a quick visual summary, the key metrics and characteristics are also illustrated in Fig. \ref{fig:radar_chart}.

For metrics related to 2D building footprints—including $\text{AP}_{50}^{\text{poly}}$, $\text{AR}_{50}^{\text{poly}}$, N-ratio, and IoU—our building polygon dataset GBA.Polygon consistently achieves the best performance across most regions, with the exception of South American cities. This demonstrates the effectiveness of our proposed quality-guided polygon fusion strategy in leveraging existing multi-source vectorized building footprint datasets.

In terms of building height-related metrics, both our raster-format height product GBA.Height and vector-format LoD1 model GBA.LoD1 achieve the highest completeness scores across all categories. This suggests that the generated height map provides the most comprehensive and meaningful coverage of building heights over the reference polygons.
For building volume estimation, as measured by $\text{RMSE}_{BV}$ and $\text{MAE}_{BV}$, our GBA.LoD1 outperforms all other compared methods across every continent except South America, confirming its suitability for city-scale volume estimation.
Lastly, in terms of building height accuracy—measured by $\text{RMSE}_{BH}$ and $\text{MAE}_{BH}$—our GBA.LoD1 delivers the best results in North America and Oceania, and remains competitive or only slightly behind other methods in the remaining regions, confirming its consistent high-quality across the globe.

Due to the lack of reference data in Africa, the results could not be directly evaluated. However, the satisfactory performance observed in South America---where both training and test data are similarly limited and building morphology and regional characteristics differ significantly from those of more developed regions---suggests that the results in Africa would likely be acceptable.

\setlength\tabcolsep{0pt}
\begin{table}
\footnotesize
\caption{
Per-continent evaluation results for various global building height products using the test dataset.
"AS", "EU", "NA", "OC", and "SA" represent Asia, Europe, North America, Oceania, and South America, respectively.
The products are grouped based on whether they provide vectorized building footprints.
Metrics of the building volume ($BV$) are reported in the units of $\text{m}^3/100~\text{m}^2$; metrics of the building heights ($BH$) are reported in m.
}
\renewcommand\arraystretch{0.85}
\newcolumntype{C}{>{\centering \arraybackslash}m{1.2cm}}
\newcolumntype{D}{>{\centering \arraybackslash}m{2.4cm}}
\newcolumntype{E}{>{\centering \arraybackslash}m{1.3cm}}
\newcolumntype{F}{>{\centering \arraybackslash}m{1.4cm}}
\begin{tabular}{C D F E E E E E E F F E}
\toprule
        Continent & Dataset & Vector & AP$_{50}^{\text{poly}}$ & AR$_{50}^{\text{poly}}$ & N-ratio &
        IoU & RMSE$_{BV}$ & MAE$_{BV}$ & RMSE$_{BH}$ & MAE$_{BH}$ & Comp. \\
        
\addlinespace[0.3ex]
\midrule
\addlinespace[0.3ex]

{\multirow{8}{*}{AS}}
 & GHS-BUILT-H & \ding{55} & - & - & - & 13.3 & 322.6 & 113.5 & \textbf{5.7} & \textbf{3.8} & 0.95 \\
 & WSF 3D & \ding{55} & - & - & - & \textbf{16.4} & 268.9 & 52.6 & 6.1 & 4.4 & 0.98 \\
 & Google 2.5D & \ding{55} & - & - & - & - & - & - & - & - & - \\
 & GBA.Height (ours) & \ding{55} & - & - & - & - & \textbf{247.7} & \textbf{41.4} & 5.8 & 4.4 & \textbf{0.99} \\
\cline{2-12}
 & Microsoft & \checkmark & 6.5 & 17.8 & 0.53 & 23.4 & - & - & - & - & - \\
 & 3D-GloBFP & \checkmark & 8.7 & 24.0 & 0.76 & 32.0 & 321.7 & 49.7 & 6.0 & \textbf{3.2} & 0.67 \\
 & GBA.LoD1 (ours) & \checkmark & \textbf{27.9} & \textbf{46.8} & \textbf{1.04} & \textbf{44.7} & \textbf{235.0} & \textbf{41.3} & \textbf{5.9} & 4.6 & \textbf{0.76} \\

\addlinespace[0.3ex]
\midrule
\midrule
\addlinespace[0.3ex]

{\multirow{8}{*}{EU}}
 & GHS-BUILT-H & \ding{55} & - & - & - & 10.2 & 206.4 & 60.4 & 7.1 & 5.0 & 0.74 \\
 & WSF 3D & \ding{55} & - & - & - & 
 \textbf{12.8} & 198.6 & 40.0 & \textbf{5.6} & 3.7 & 0.96 \\
 & Google 2.5D & \ding{55} & - & - & - & - & - & - & - & - & - \\
 & GBA.Height (ours) & \ding{55} & - & - & - & - &  
 \textbf{150.9} & \textbf{24.3} & \textbf{5.6} & \textbf{3.5} & \textbf{0.99} \\
\cline{2-12}
 & Microsoft & \checkmark & 13.9 & 28.1 & 0.65 & 52.6 & 168.4 & 27.4 & \textbf{3.9} & \textbf{2.8} & 0.55 \\
 & 3D-GloBFP & \checkmark & 5.1 & 12.0 & 0.31 & 25.1 & 257.1 & 47.2 & 6.2 & 4.6 & 0.52 \\
 & GBA.LoD1 (ours) & \checkmark & \textbf{46.0} & \textbf{69.3} & \textbf{1.19} & \textbf{67.2} & \textbf{139.4} & \textbf{22.7} & 4.1 & 3.0 & \textbf{0.83} \\

\addlinespace[0.3ex]
\midrule
\midrule
\addlinespace[0.3ex]

{\multirow{8}{*}{NA}}
 & GHS-BUILT-H & \ding{55} & - & - & - & 10.3 & 185.1 & 44.3 & 8.8 & 4.0 & 0.93 \\
 & WSF 3D & \ding{55} & - & - & - & \textbf{12.7} & 173.4 & 23.8 & \textbf{5.9} & 2.9 & \textbf{0.99} \\
 & Google 2.5D & \ding{55} & - & - & - & - & - & - & - & - & - \\
 & GBA.Height (ours) & \ding{55} & - & - & - & - & \textbf{144.1} & \textbf{16.6} & 6.4 & \textbf{2.4} & \textbf{0.99} \\
\cline{2-12}
 & Microsoft & \checkmark & 37.8 & 58.0 & 0.94 & 58.5 & 149.3 & 16.1 & 6.4 & 2.6 & 0.81 \\
 & 3D-GloBFP & \checkmark & 41.3 & 55.5 & 0.77 & 63.2 & 153.9 & 18.3 & 8.2 & 6.7 & 0.90 \\
 & GBA.LoD1 (ours) & \checkmark & \textbf{54.1} & \textbf{72.4} & \textbf{1.09} & \textbf{66.8} & \textbf{135.5} & \textbf{15.1} & \textbf{5.3} & \textbf{2.1} & \textbf{0.96} \\

\addlinespace[0.3ex]
\midrule
\midrule
\addlinespace[0.3ex]

{\multirow{8}{*}{SA}}
 & GHS-BUILT-H & \ding{55} & - & - & - & 25.1 & 560.2 & 265.4 & 11.3 & 6.5 & 0.98 \\
 & WSF 3D & \ding{55} & - & - & - & 29.4 & 566.6 & 198.7 & 13.8 & 7.4 & 0.99 \\
 & Google 2.5D & \ding{55} & - & - & - & \textbf{46.6} & \textbf{519.7} & 188.4 & \textbf{8.6} & \textbf{5.1} & 0.97 \\
 & GBA.Height (ours) & \ding{55} & - & - & - & - & 580.0 & \textbf{171.5} & 10.1 & 5.7 & \textbf{1.00} \\
 \cline{2-12}
 & Microsoft & \checkmark & 0.0 & 0.6 & 0.10 & 42.2 & - & - & - & - & - \\
 & 3D-GloBFP & \checkmark & 0.1 & 1.2 & 0.14 & \textbf{43.4} & 1708.5 & 697.3 & 29.2 & 24.5 & \textbf{0.89} \\
 & GBA.LoD1 (ours) & \checkmark & \textbf{0.5} & \textbf{4.8} & \textbf{0.69} & 37.6 & \textbf{586.8} & \textbf{180.6} & \textbf{8.9} & \textbf{5.0} & 0.87 \\

\addlinespace[0.3ex]
\midrule
\midrule
\addlinespace[0.3ex]

{\multirow{8}{*}{OC}}
 & GHS-BUILT-H & \ding{55} & - & - & - & 3.9 & 61.4 & 11.6 & 1.9 & 1.4 & 0.85 \\
 & WSF 3D & \ding{55} & - & - & - & \textbf{8.6} & 42.7 & 3.9 & 2.1 & 1.4 & 0.97 \\
 & Google 2.5D & \ding{55} & - & - & - & - & - & - & - & - & - \\
 & GBA.Height (ours) & \ding{55} & - & - & - & - & \textbf{38.0} & \textbf{3.6} & \textbf{1.5} & \textbf{1.0} & \textbf{0.99} \\
\cline{2-12}
 & Microsoft & \checkmark & 18.9 & 46.1 & 1.15 & 37.5 & 46.7 & \textbf{3.9} & \textbf{1.5} & \textbf{1.0} & 0.57 \\
 & 3D-GloBFP & \checkmark & 19.2 & 45.9 & \textbf{1.14} & 38.7 & \textbf{52.2} & 4.8 & 1.8 & 1.1 & 0.84 \\
 & GBA.LoD1 (ours) & \checkmark & \textbf{36.7} & \textbf{56.0} & 1.16 & \textbf{39.8} & 46.8 & 4.2 & \textbf{1.5} & \textbf{1.0} & \textbf{0.87} \\

\bottomrule
\end{tabular}
\label{tab:comparison_results}
\end{table}

\subsubsection{Qualitative Results}

\begin{figure}[H]
    %\ContinuedFloat
    %\clearpage 
    \centering
    \includegraphics[width=1.0\textwidth]    {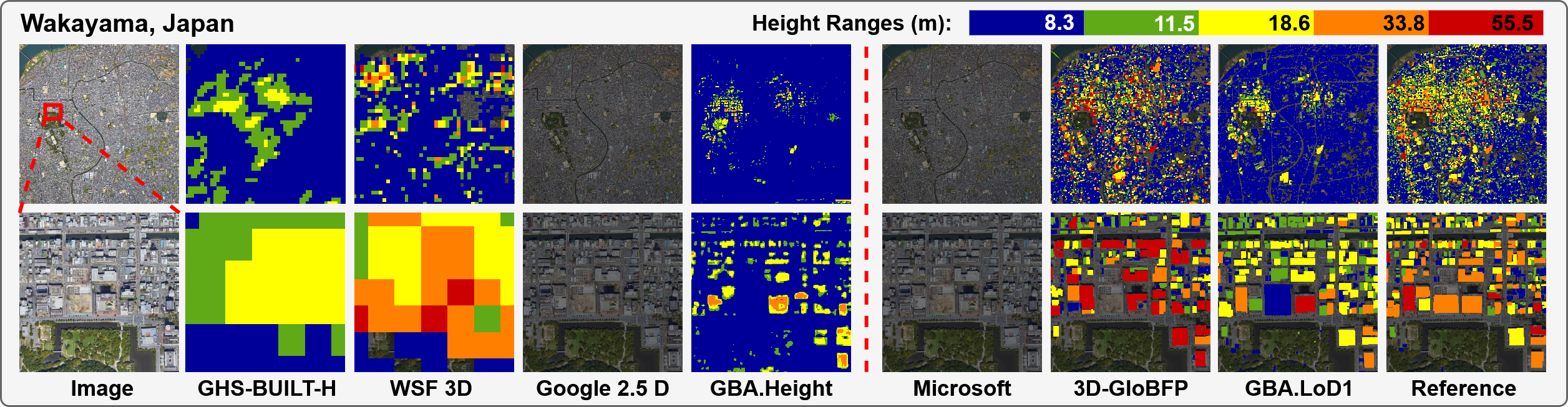}\vspace*{-5pt}
   \noindent (e) Wakayama, Japan, Asia
    %\includegraphics[width=1.0\textwidth]{figures/visual_comparison_v5_part1.pdf}
    %\caption{Visual comparison of existing building height products on test cities including Portland (North America), Medellín (South Ameirca), and Bordeaux (Europe). (a)}
    %\label{fig:visual_comparison_combined_a}

    %\includegraphics[width=1.0\textwidth]{figures/visual_comparison_v5_part2.pdf}
    \caption{Visual comparison of existing building height products on the test cities Portland (North America), Medellín (South America), Bordeaux (Europe), Launceston (Oceania) and Wakayama (Asia).}
    \label{fig:visual_comparison_combined}
\end{figure}

To provide an intuitive understanding of the performance of various large-scale building height products, we visualize the results from the different methods on selected test cities across multiple continents: Portland, USA for North America, Medellín, Colombia for South America, Bordeaux, France for Europe, Launceston, Australia for Oceania, and Wakayama, Japan for Asia.
As shown in Fig.  \ref{fig:visual_comparison_combined}, the raster-based products, such as GHS-BUILT-H (\cite{Pesaresi2024,Pesaresi2023}) and WSF 3D (\cite{Esch2020,Esch2022}), provide coarse, city-scale building height estimates but fail to capture the detailed building geometries---a critical limitation for many downstream applications.
Google 2.5D (\cite{Sirko2023}) produces accurate building heights in South American cities, like Medellín, but its geographic coverage is limited primarily to the Global South.
Microsoft’s LoD1 model offers detailed polygon-level height estimations with good spatial resolution. However, it lacks coverage in Asia and South America and tends to underestimate heights in European and Asian cities, such as Bordeaux and Wakayama.
3D-GloBFP (\cite{essd-16-5357-2024}) provides global but incomplete coverage, yet tends to overestimate building heights, especially in Medellín and Bordeaux.
In contrast, our building height map GBA.Height and LoD1 model GBA.LoD1 achieve accurate estimations in cities across North America, Europe, and Oceania, including Portland, Bordeaux, and Launceston.
While some underestimation is still observed in Medellín and Wakayama, our models strike a significantly better overall balance between resolution, completeness, and accuracy compared to existing solutions.

\subsection{Validation of the UN's Building Count Estimation}
The UN has estimated that there are approximately 4 billion buildings worldwide (\cite{unhabitat2019}), a figure that exceeds the number of building instances in any existing datasets, including our GBA. However, by using the N-ratio evaluation metric shown in Table \ref{tab:comparison_results}, which reflects the completeness of GBA.Polygon, we provide an alternative estimate of the global building count based on our product to validate the UN's estimate.

Initially, N-ratios were computed continentally and globally against reference data, and ranged from 0.69 in South America to 1.19 in Europe, with a global average of 1.03. Using these ratios, we estimated the total number of buildings per continent by scaling the known building counts accordingly. For Africa, where no N-ratio is available, we computed three estimates using the minimum, maximum, and global average N-ratios from other continents.

Aggregate projections across continents yield a global estimate of 2.71 billion buildings, while the estimated bounds range from 2.64 billion to 2.97 billion buildings. This analysis suggests that the UN's estimate may exceed actual global building counts.

\begin{figure}[H]  % Use [p] for page float
    \centering
    \includegraphics[width=1.0\textwidth]{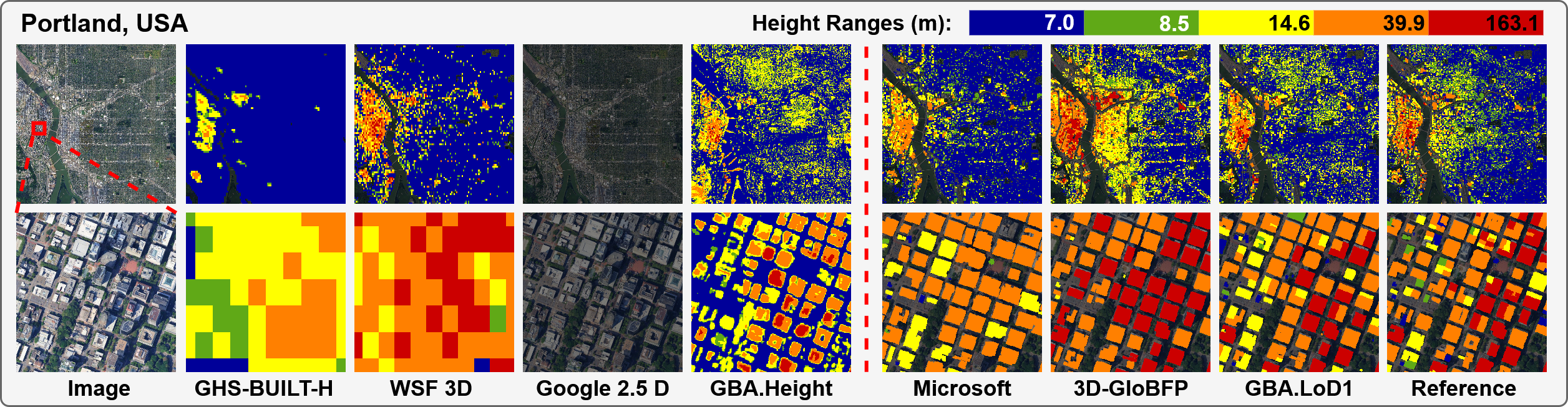}\vspace*{-5pt}
   \noindent (a) Portland, USA, North America
   
    \vspace{9pt}
    \includegraphics[width=1.0\textwidth]{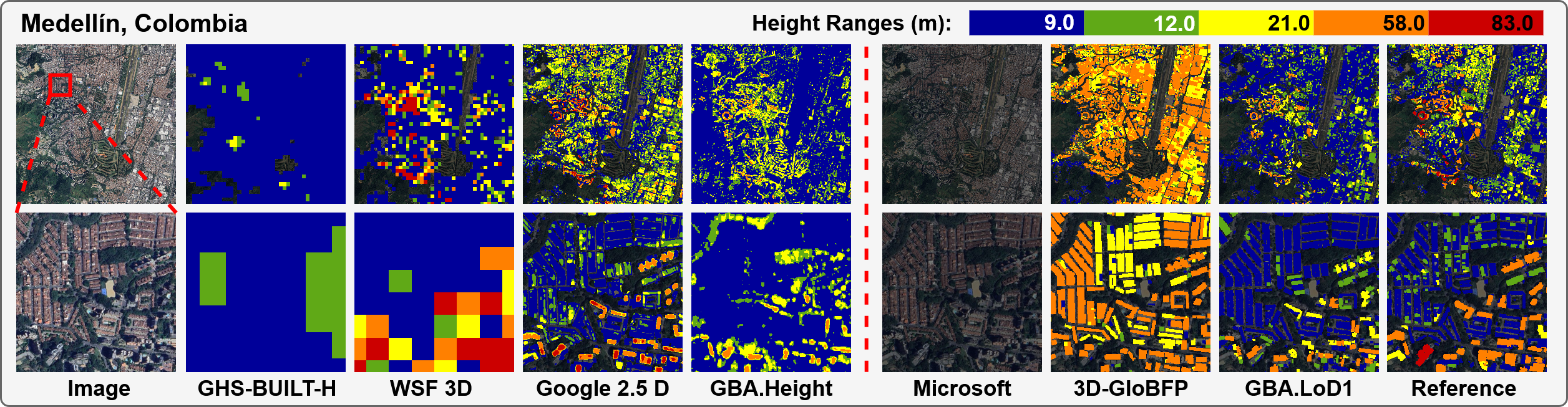}\vspace*{-5pt}
   \noindent (b) Medell\'{i}n, Colombia, South America

   \vspace{9pt}
    \includegraphics[width=1.0\textwidth]{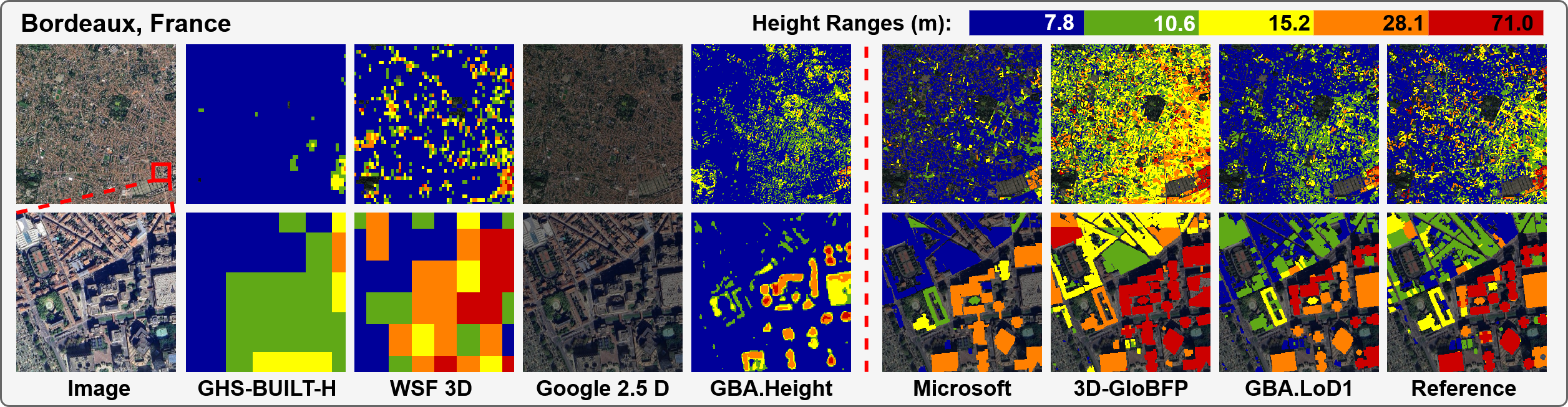}\vspace*{-5pt}
   \noindent (c) Bordeaux, France, Europe
   
   \vspace{9pt}
    \includegraphics[width=1.0\textwidth]{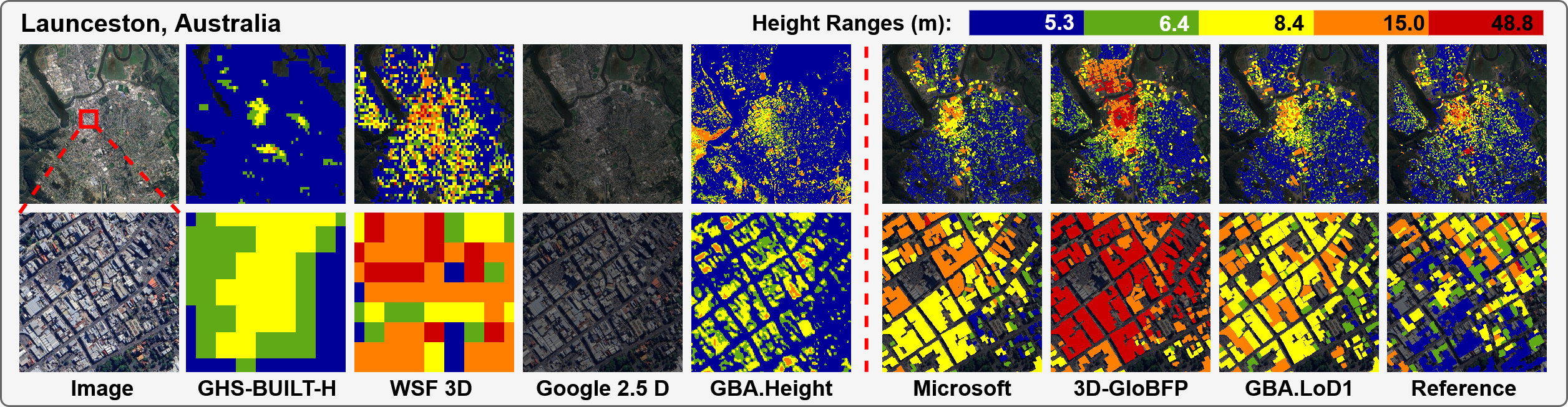}\vspace*{-5pt}
   \noindent (d) Launceston, Australia, Oceania 
    
\end{figure}

\afterpage{\clearpage}

\begin{comment}
\iffalse
\begin{figure}[htp]  % Allow page breaks and continued float
    \ContinuedFloat  % Tell LaTeX this is a continuation of previous figure
    \centering
    \includegraphics[width=1.0\textwidth]{figures/fig4_1.png}
    \includegraphics[width=1.0\textwidth]{figures/fig4_2.png}
    \includegraphics[width=1.0\textwidth]{figures/fig4_3.png}
    \includegraphics[width=1.0\textwidth]{figures/fig4_4.png}
\end{figure}

\begin{figure}[htp]
    \ContinuedFloat
    \includegraphics[width=1.0\textwidth]{figures/fig4_5.png}
    \caption{ALTERNATIVE: Visual comparison of existing building height products on test cities including Launceston (Oceania) and Wakayama (Asia). (b)}
    %\label{fig:visual_comparison_combined_b}
\end{figure}

\fi
\end{comment}

\subsection{Discussions}
Based on the preceding analysis of our dataset, its key strengths and limitations can be summarized as follows.

\subsubsection*{Strengths}
\begin{itemize}
    \item GBA.Polygon represents the most comprehensive dataset of global building polygons to date, comprising approximately 2.75 billion buildings. This achievement could be primarily attributed to two key factors: (1) the development of a multi-source building polygon fusion strategy that effectively integrates all existing large-scale polygon datasets; and (2) the implementation of a global building polygon generation pipeline based on PlanetScope satellite imagery, which addresses existing data gaps by producing previously unavailable building footprints, albeit with limited quality in some regions constrained by a lower spatial resolution of the input satellite data.
    
  \item GBA.Height represents the first high-resolution building height map available at the global scale. In contrast to existing global building height datasets~(\cite{Pesaresi2023,Ma2024,Esch2022}), which offer coarse spatial resolutions ranging from 90 to 150 meters, our approach can achieve a resolution of 3 meters—representing at least a 30-fold improvement. This substantial enhancement enables more detailed and accurate representation of urban structures, facilitating a broader range of downstream geospatial applications. Furthermore, height estimation in GBA.Height is derived exclusively from a single and affordable data source, namely PlanetScope satellite imagery, instead of very high-resolution aerial data. Owing to the reduced data dependency data cost and the high temporal revisit frequency of PlanetScope, our approach supports rapid and scalable updates of global building height maps. 

 \item GBA.LoD1 constitutes the most comprehensive global LoD1 building model to date, encompassing approximately 2.68 billion buildings and filling a gap of over 1 billion previously unrepresented structures worldwide. Compared to existing LoD1 datasets, GBA.LoD1 offers globally consistent and high-accuracy building height estimates, with RMSEs ranging from 1.5 to 8.9 meters across different geographic regions.
    
\end{itemize}

\subsubsection*{Limitations}
\begin{itemize}
    \item Limited availability of height data in Africa for training and validation. Due to the scarcity of accurate reference data in Africa, our building height estimation model was neither trained nor validated on samples from this region. Consequently, the model may be subject to domain shift, potentially limiting its generalizability and accuracy when applied to urban areas in African contexts. The limitations in available training data could be potentially addressed by applying weakly supervised learning, i.e. training dedicated models by considering all existing building height data as weak labels. However, community efforts will still be needed to curate the validation dataset.
        
    \item Underestimation of the heights in high-rise building areas.
As shown in the visual results in Fig. \ref{fig:visual_comparison_combined}, our monocular height estimation model tends to underestimate building heights in some South American and Asian cities. This may be attributed to the model being trained to minimize pixel-level height errors rather than building instance-level errors, leading to more conservative predictions in unfamiliar high-rise urban scenarios.

\end{itemize}

%%%%%%%%%%%%%%%%%%%%%%%%%%%%%%%%%%%%%%%%%%%%%
\section{Applications and Enrichment}\label{useCases}
In this section, we demonstrate two applications of the global LoD1 building models. The first examines the correlation between building volume and population, while the second evaluates the utility of building volume information for computing SDG indicators.

\subsection{Correlation with the Population}
\subsubsection{Fine-scale Continental Analysis: Europe as a Case Study}
\begin{figure}
    \centering
    \includegraphics[width=\linewidth]{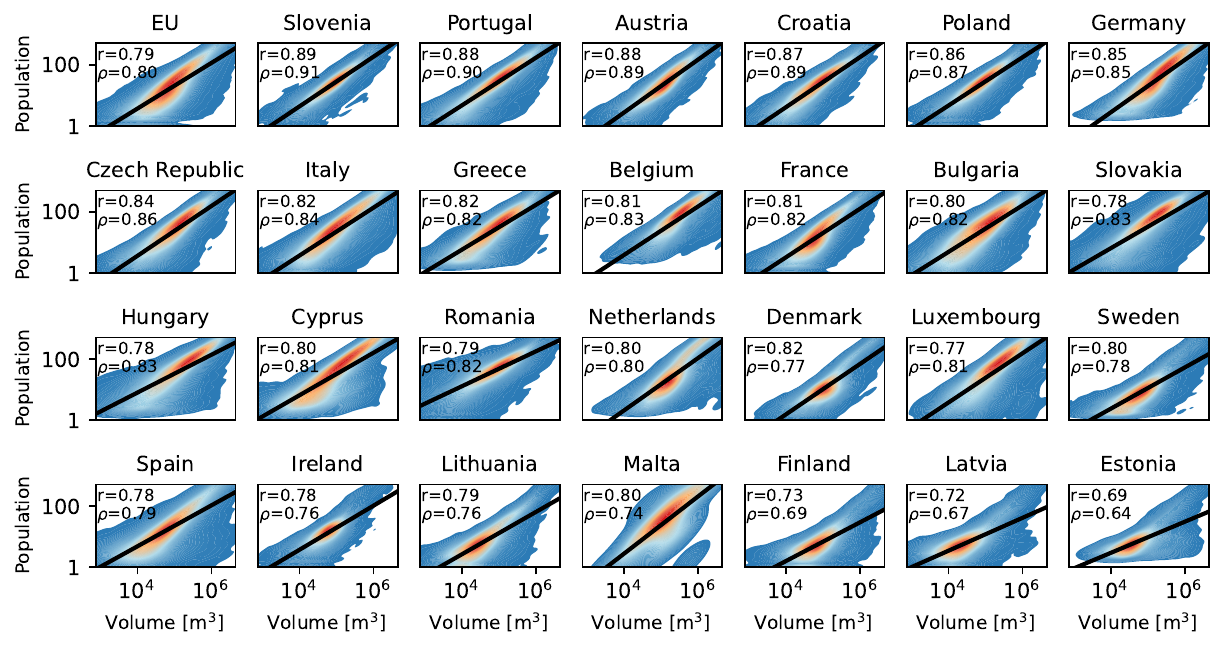}
    \caption{Regression analysis between population and building volume were conducted for the entire EU as well as for each of the 27 EU member states individually. The first graph shows the regression analysis of the EU as a whole, followed by the 27 member countries sorted in descending order based on the harmonized metric that averages the Pearson ($r$) and Spearman ($\rho$) correlation coefficients. (Year 2021)}
    \label{fig:eu_pop}
\end{figure}

We investigated the relationship between population distribution and building volume across a 1 km $\times$ 1 km grid over the 27 member states of the European Union (EU). The population data used in this analysis were sourced from the EU's official population grid dataset in year 2021 (\cite{eurostat_geostat}). As can be seen in Fig. \ref{fig:eu_pop}, we performed a logarithmic regression between the population count and building volume for each grid cell, both at the EU-wide level and for individual member states. To assess the significance of the relationship, we computed both the Pearson correlation coefficient ($r$) and the Spearman rank correlation coefficient ($\rho$), which measure linear and monotonic association, respectively. 

In general, population exhibits a strong positive correlation with building volume: areas with higher populations tend to be associated with buildings of greater volume. While the slope of the regression line reflects the average building volume per capita, the correlation coefficient provides insights into the equity of the distribution. Higher correlation values suggest a more uniform allocation of building volume relative to population across spatial units, indicating a closer alignment between the built infrastructure and population density.

\begin{figure}
    \centering
    \includegraphics[width=\linewidth]{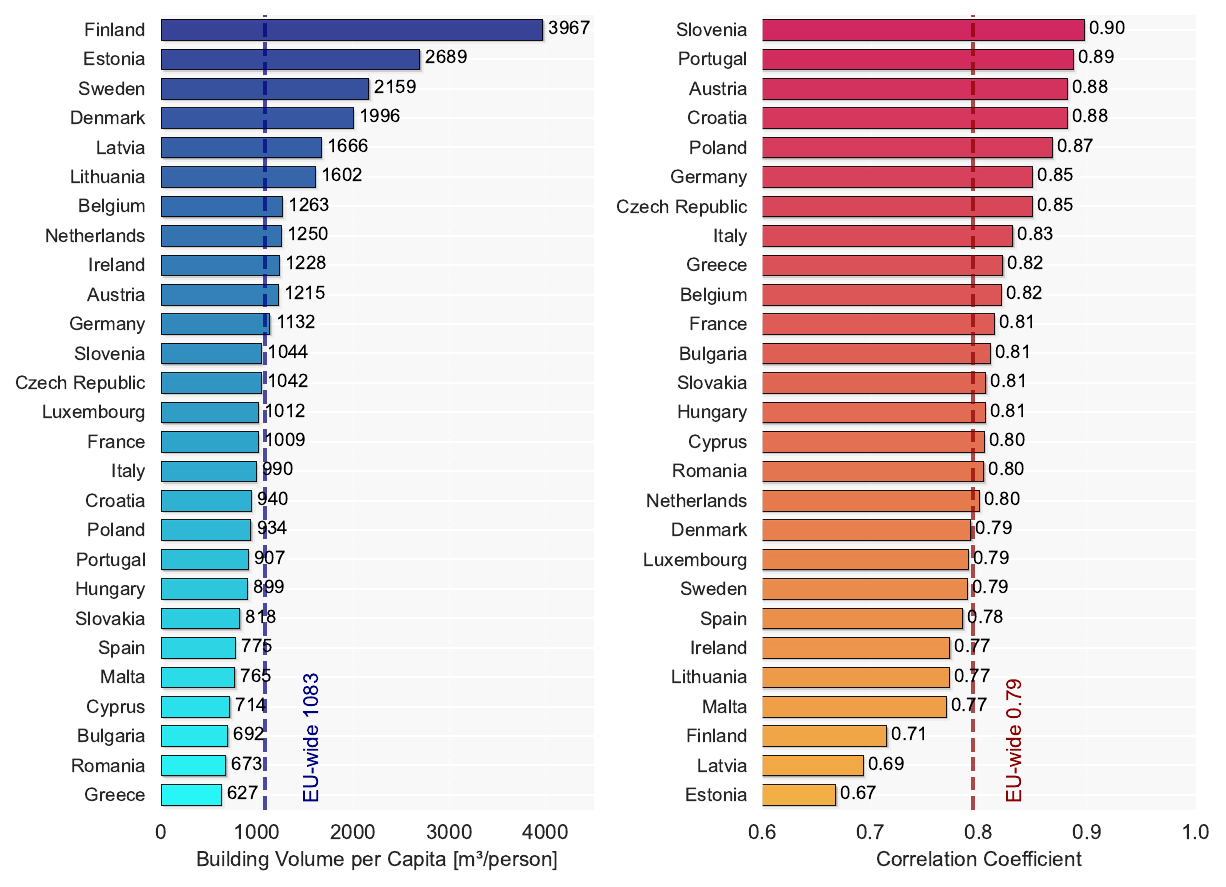}
    
    \caption{Building volume per capita and the harmonized correlation coefficients for the 27 EU member states and the EU as a whole. The harmonized correlation coefficients were computed as an average of the Pearson ($r$) and Spearman ($\rho$) correlation coefficients. (Year 2021)}
    \label{fig:eu_bar}
\end{figure}

As illustrated in Fig. \ref{fig:eu_bar}, a considerable disparity emerges when comparing the building volume per capita and correlation coefficients across the 27 EU countries. More than half of the member states fall below the EU-wide average in terms of building volume per capita. Finland, which ranks highest in this metric, possesses six times the per-capita building volume of Greece, which ranks lowest. The disparities are somewhat less pronounced in the correlation coefficients, with most countries exhibiting relatively strong correlations, with only seven countries falling significantly below the EU-wide correlation benchmark.

\begin{comment}
\iffalse
\clearpage
\begin{figure}
    \ContinuedFloat
    \centering
    
    \includegraphics[width=\linewidth]{figures/fig7_YW.pdf}
    \caption{ALTERNATIVE COLOR. Building volume per capita and the harmonized correlation coefficient for the 27 EU member states and the EU as a whole. The harmonized correlation coefficient is computed as an average of the Pearson ($r$) and Spearman ($\rho$) correlation coefficients. (Year 2021)}
    \label{fig:eu_bar}
\end{figure}
\clearpage
\begin{figure}
    \ContinuedFloat
    \centering
    
    \includegraphics[width=\linewidth]{figures/fig7_YW_v2.pdf}
    \caption{ALTERNATIVE PLOT. Building volume per capita and the harmonized correlation coefficient for the 27 EU member states and the EU as a whole. The harmonized correlation coefficient is computed as an average of the Pearson ($r$) and Spearman ($\rho$) correlation coefficients. (Year 2021)}
    \label{fig:eu_bar}
\end{figure}

\fi
\end{comment}

Notably, some of the countries with the highest building volume per capita, such as Finland and Estonia, exhibit a particularly imbalanced allocation of building volume with respect to their population, as evidenced by their lower correlation coefficients. This suggests that while they have abundant building stock, its distribution is less aligned with their population density.
\subsubsection{Coarse-scale Global Analysis}
\begin{figure}
    \centering
    \includegraphics[width=0.88\linewidth]{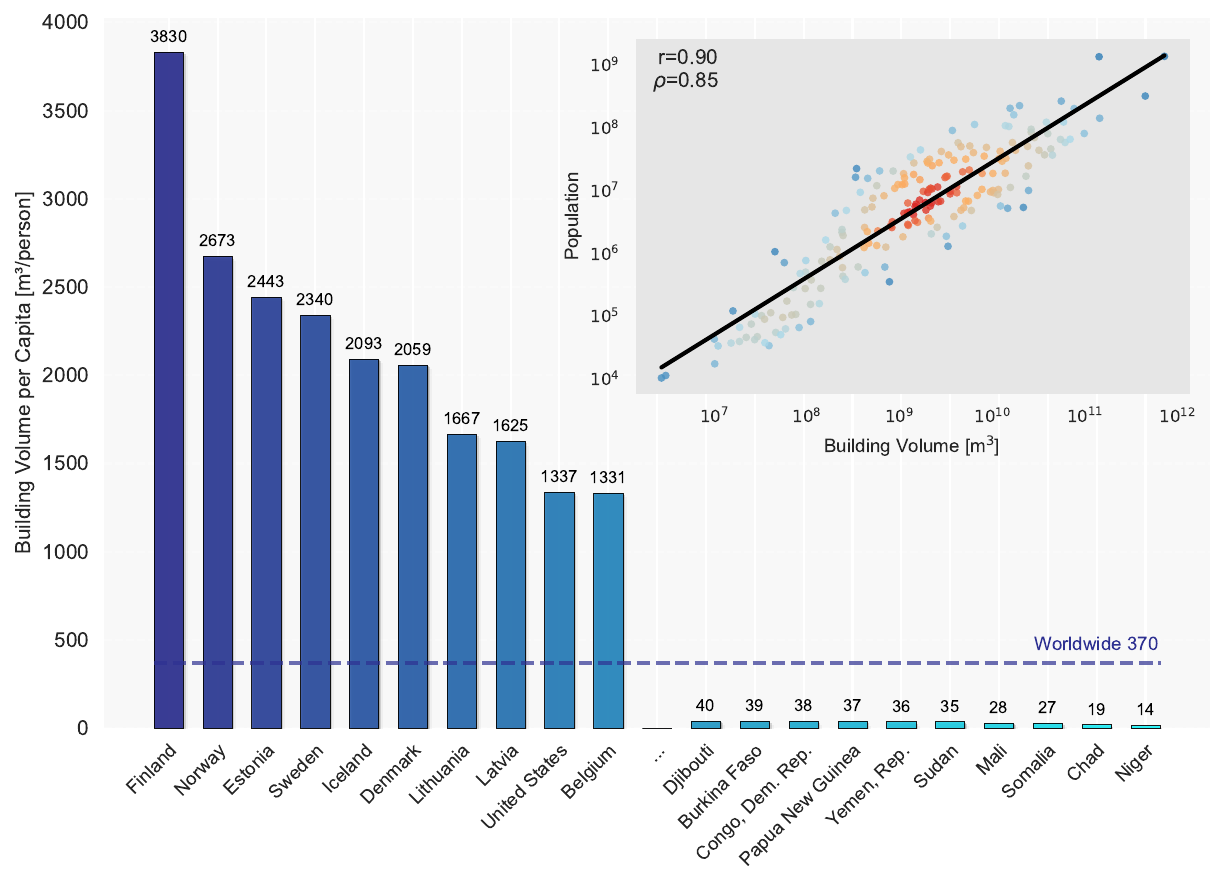}
    \caption{Regression analysis of population and building volume across all countries and territories globally with Pearson (r) and Spearman (ρ) correlation coefficients. Top 10 and bottom 10 countries or territories are displayed by building volume per capita. (Year 2019)}
    \label{fig:global_pop}
\end{figure}

Similar conclusions can be drawn from a coarse, country-level analysis conducted at the global scale. We performed a logarithmic regression between the total population of each country or territory (\cite{worldbank_population}) in year 2019 and the corresponding total building volume derived from our dataset. As can be seen in Fig. \ref{fig:global_pop}, the regression again indicates a strong positive correlation between the population and building volume. 

However, the analysis also reveals a much greater imbalance in building volume per capita across countries. Countries with the largest building volumes per capita are predominantly located in Europe, while those with the smallest values are primarily in Africa. As an example, Finland continues to exhibit the highest building volume per capita, whereas in Niger, Africa, individuals have access to only 0.36\% of that value. This figure is also 27 times lower than the global average, highlighting significant disparities in built infrastructure worldwide.

\begin{comment}
\iffalse
\clearpage
\begin{figure}
    \ContinuedFloat
    \centering
    \includegraphics[width=\linewidth]{figures/fig8_YW.pdf}
    \caption{ALTERNATIVE: Regression of population and building volume across all countries and territories globally with Pearson (r) and Spearman (ρ) correlation coefficients. Top 10 and bottom 10 countries or territories ranked by building volume per capita. (Year 2019)}
\end{figure}
\fi
\end{comment}

\subsection{Potential SDG Indicator}
To monitor progress toward SDG 11---\textit{Make cities and human settlements inclusive, safe, resilient, and sustainable}---several indicators have been defined (\cite{unsdg_indicator}). Among these, Indicator 11.3.1---\textit{Ratio of land consumption rate to population growth rate}---plays a crucial role (\cite{unhabitat_11_3_1_metadata}). %This indicator is computed by deriving changes in built-up area and population. 
The indicator is computed by deriving the built-up area change and population change. Due to the absence of time series data, we were unable to directly compute the indicator to substantiate the rationale for employing volume-based indicators, though generating time series data using our developed pipeline is relatively straightforward. As an alternative approach, we examined the use of volume-based indicators by considering the secondary indicator defined in the indicator metadata (\cite{unhabitat_11_3_1_metadata})---\textit{built-up area per capita}. In this context, we substituted built-up area with built-up volume to assess whether the revised indicator could provide a more accurate reflection of the urban development status.

Sustainable development is influenced not only by factors related to the built-up environment but also by a wide range of other social, economic, and environmental variables. Nevertheless, correlation analysis between built-up environment indicators and development-related measures can provide insights into the relative effectiveness of different indicators. In this study, we used gross domestic product (GDP) per capita in year 2019 (\cite{worldbank_gdp_per_capita}) as a proxy for development status and compared its correlation with both the building area per capita and building volume per capita. The per-capita building metrics were computed using population data for the year 2019 (\cite{worldbank_population}).

\begin{figure}
    \centering
    \includegraphics[width=0.8\linewidth]{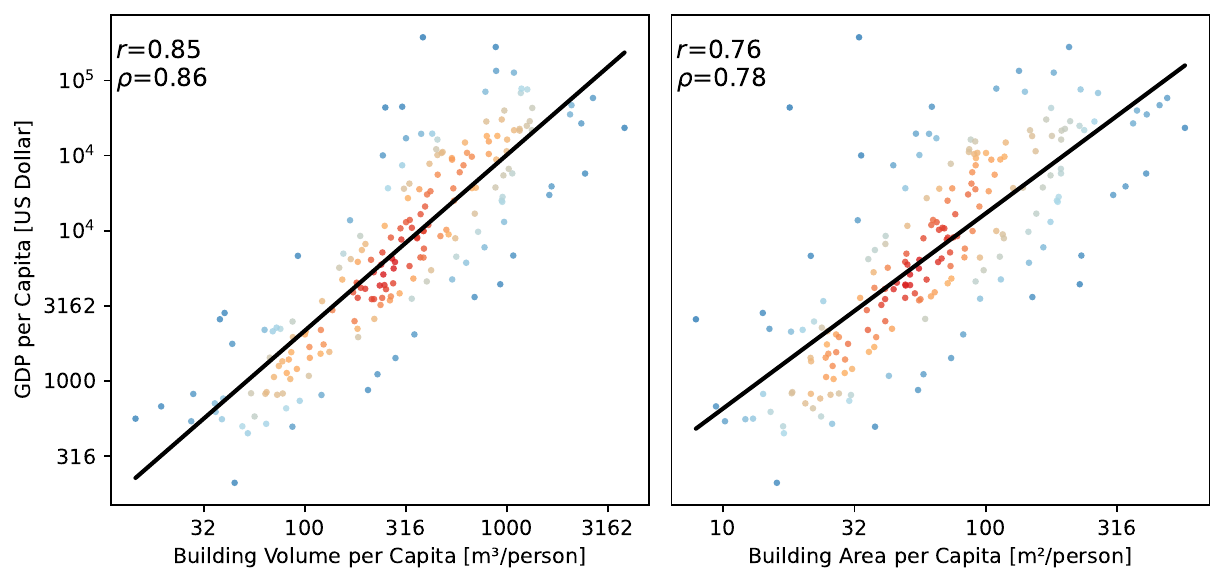}
    \caption{Correlations between GDP per capita and two built environment indicators: building volume per capita and building area per capita.  Pearson (r) and Spearman (ρ) correlation coefficients are reported. (Year 2019)} 
    \label{fig:sdg_indicator}
\end{figure}

As Fig. \ref{fig:sdg_indicator} shows, GDP per capita exhibits a strong positive correlation with both building area per capita and building volume per capita, with correlation coefficients of 0.76 and 0.85, respectively. The significantly higher correlation coefficient between building volume per capita and GDP per capita suggests building volume per capita might be a more suitable measure.

%\begin{figure}
%    \begin{subfigure}
%        \centering
%        \includegraphics[width=\linewidth]{figures/fig9.pdf}
%    \end{subfigure}
    
%    \begin{subfigure}
 %       \centering
 %       \includegraphics[width=\linewidth]{figures/fig9_YW.pdf}
 %       \caption{ALTERNATIVE: Correlations between GDP per capita and two built environment indicators: building volume per capita and building area per capita. (Year 2019)}        
 %   \end{subfigure}

 %   \label{fig:sdg_indicator}
%\end{figure}

To further validate this observation, we assessed the alignment between GDP per capita and building-based indicators—namely, building volume per capita and building area per capita—by randomly selecting pairs of countries and comparing their respective rankings. For each pair, we examined whether the ranking based on the indicator agreed with the ranking based on GDP per capita. As shown in Table \ref{tab:indicator}, among 20,301 country or territory pairs sampled from a total of 202 countries and territories, using building volume per capita instead of building area per capita resulted in 788 additional ranking agreements. This substitution increased the agreement rate from 79.6\% (area-based) to 83.5\% (volume-based), demonstrating the superior discriminative power of building volume per capita in capturing differences in economic development.

\begin{table}[h!]
\newcolumntype{C}{>{\centering \arraybackslash}m{1.2cm}}
\newcolumntype{D}{>{\centering \arraybackslash}m{2.2cm}}
\newcolumntype{E}{>{\centering \arraybackslash}m{2.7cm}}
\newcolumntype{F}{>{\centering \arraybackslash}m{3.5cm}}
\footnotesize
\centering
\caption{Comparison of the ranking accuracy for GDP per capita estimation using building volume per capita and building area per capita. (Year 2019)}
\begin{tabular}{F F F F F}
\toprule
Indicator & Agreement Between Indicators & Only One Indicator Correct & Total Correct Rankings & Ranking Accuracy [\%]  \\
\midrule
\addlinespace[0.5ex]
 Volume per Capita & 15724 & \textbf{1229} & \textbf{16953} & \textbf{83.5} \\
 Area per Capita & 15724 & 441 & 16165 & 79.6 \\ 
\bottomrule
\end{tabular}
\label{tab:indicator}
\end{table}

In conclusion, volume-based indicators demonstrate greater potential than area-based metrics for assessing progress toward the SDGs related to sustainable urban development.
%%%%%%%%%%%%%%%%%%%%%%%%%%%%%%%%%%%%%%%%
\section{Code and Data Availability}\label{dataaccess}
All code is available on GitHub under an MIT license with the Commons Clause: \url{https://github.com/zhu-xlab/GlobalBuildingAtlas} (last
access: 31 May 2025). The Commons Clause restricts the use of this software for commercial purposes. The repository includes the full development code for the GBA product, as well as scripts for reproducing the figures in this manuscript. A portal for interactive preview of the dataset is available at the same URL.

The GBA dataset described in this manuscript can be accessed on mediaTUM under \url{https://mediatum.ub.tum.de/1782307} (\cite{mediatum}).
%%%%%%%%%%%%%%%%%%%%%%%%%%%%%%%%%%%%%%%%
\section{Conclusions}\label{conclusions}
In this study, we introduce GlobalBuildingAtlas, an open global dataset comprising building polygons, heights, and LoD-1 3D building models. Our motivation stems from wanting to fill the significant gaps in existing datasets, particularly regarding the lack of comprehensive 3D information at an individual building level, which limits applications in both large-scale urban environments and detailed assessments.

We developed a streamlined pipeline that utilizes only optical satellite imagery for LoD-1 3D building modeling. This approach facilitates rapid updates for time-series monitoring and enables a wide range of applications. The pipeline consists of two main components: building polygon generation and building height estimation. The building polygon generation process addresses gaps in current building footprint datasets by supplementing missing building instances. Building height estimation produces pixel-wise height maps at a resolution of 3 m, extending applications beyond the built environment alone. The resulting LoD1 3D building model constitutes the most comprehensive and accurate dataset to date, encompassing 2.75 billion buildings, with 97.7\% of them having height estimates. Our height estimation achieves high accuracy, with RMSEs ranging from 1.5 m to 8.9 m. Additionally, the resulting building volumes demonstrates significant advantages over existing products, with RMSEs ranging from 46.8 $\text{m}^3/100~\text{m}^2$ to 586.8 $\text{m}^3/100~\text{m}^2$.

We further explored the relationship between population and building volume, revealing a strong correlation. Our analysis highlighted the substantial disparities in building volume per capita on both continental and global scales. Moreover, we propose using building volume-based indicators to assess progress toward the SDGs, as these indicators better reflected development status compared to building area-based metrics in our analysis.
\clearpage
\appendix
\section{Building Volume Distribution by Country or Territory}
\begin{figure}[ht]
    \centering
    \includegraphics[width=0.9\linewidth,trim={0 0 0 11cm},clip]{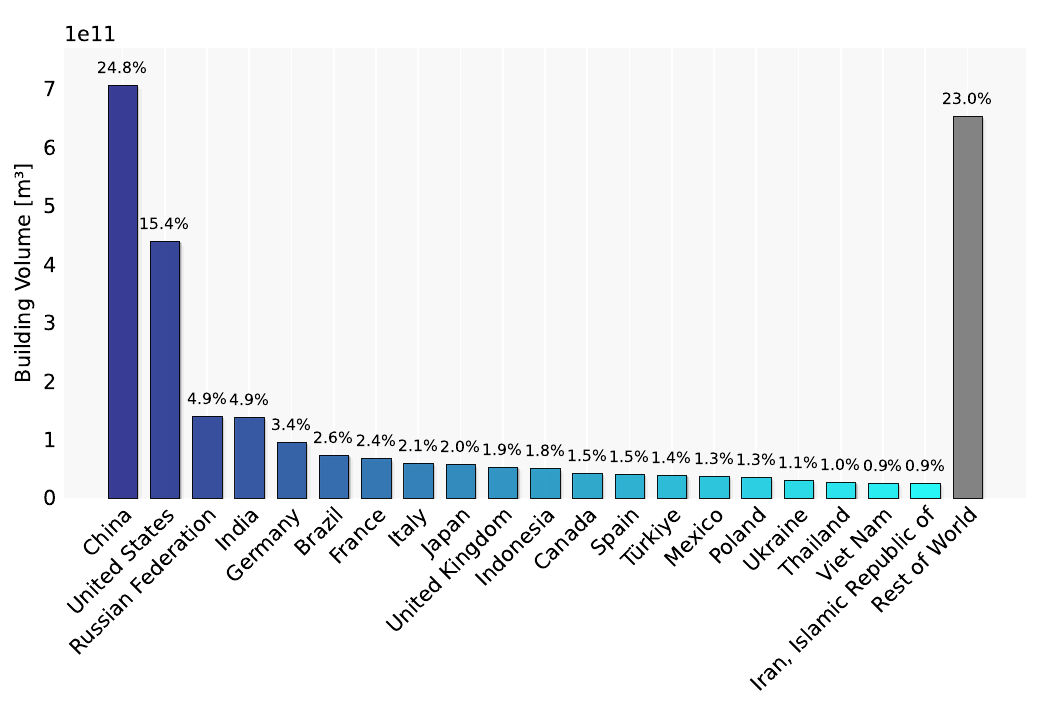}
    \caption{Building volume distribution by country or territory. Top: spatial distribution of building volumes; Bottom: Top 20 countries or territories with most building volumes. The colorbars are consistent.}
    \label{fig:volume_dist}
\end{figure}

Figure \ref{fig:volume_dist} illustrates the distribution of building volumes by country or territory. The results reveal that China holds the largest global share, accounting for 24.8\% of total building volume, followed by the USA at 15.4\%. These findings are consistent with the prior study (\cite{essd-16-5357-2024}). Notably, due to improved building data completeness, higher proportions of building volumes for Russia, India, and Brazil are uncovered. Considering the high population count and land area in these countries and territories, the finding seems to be reasonable. Conversely, building volumes for other countries and territories are estimated to constitute a smaller percentage compared to the earlier research, which aligns more closely with the expected patterns.
\begin{comment}
\iffalse
\begin{figure}[ht]
    \centering
    \includegraphics[width=0.8\linewidth]{figures/combined_plots_volume_by_country.pdf}
    \caption{Building volume distribution by country or territory. Top: spatial distribution of building volumes; Bottom: Top 20 countries or territories with most building volumes. The colorbars are consistent.}
    \label{fig:volume_dist}
\end{figure}
\clearpage
\fi
\end{comment}

\section{Contribution of Different Data Sources in Quality-guided Building Polygon Fusion}

\begin{figure}[ht]
    \centering
    \includegraphics[width=\linewidth]{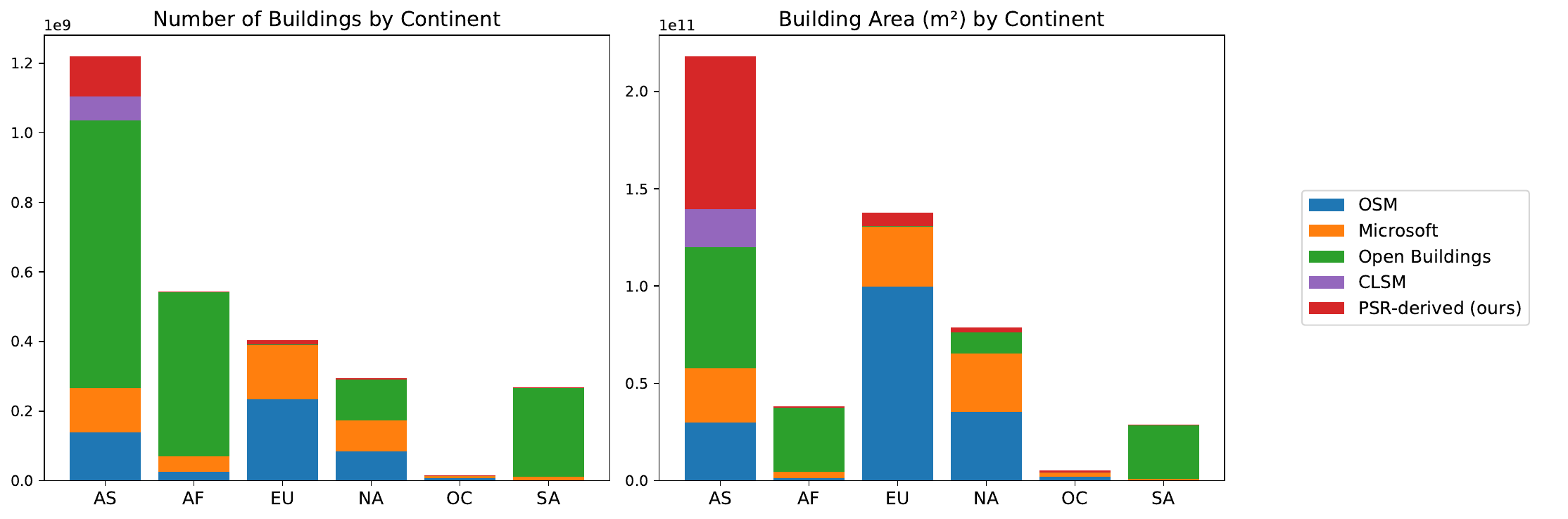}
    \caption{Contribution of various building polygon sources by continent in the quality-guided polygon fusion process. The number of buildings and total building area are attributed to each source. “PSR-derived (ours)” refers to the building footprint polygons generated from PSR as described in Sect. \ref{sec:polygon_generation}.}
    \label{fig:contribution_bar}
\end{figure}

\begin{figure}[ht]
    \centering
    \includegraphics[width=1.0\linewidth]{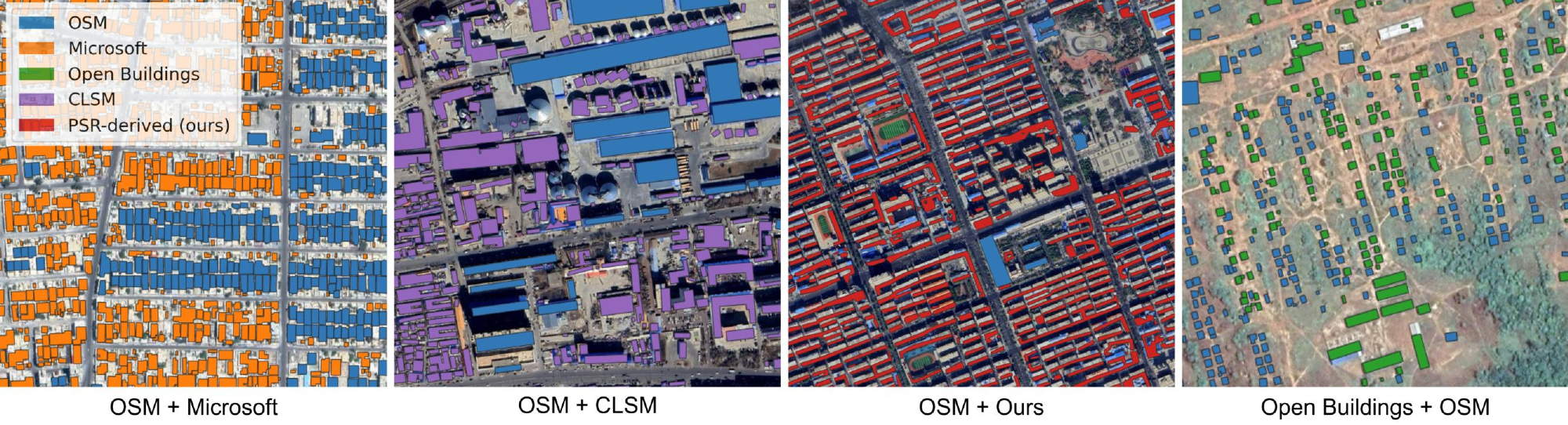}
    \caption{Examples of fused building polygons derived from multiple footprint sources. “PSR-derived (ours)” refers to the building footprint polygons generated from PSR as described in Sect. \ref{sec:polygon_generation}.}
    \label{fig:contribution_sample}
\end{figure}

Figure~\ref{fig:contribution_bar} illustrates the contributions of various building data sources to GBA.Polygon.
In terms of building count, Open Buildings~(\cite{GoogleOpenBuildings}) is the largest contributor in Asia, Africa, and South America, with a total of 1.62 billion buildings.
OpenStreetMap (OSM) provides 0.49 billion buildings, showing more complete coverage in Europe and North America, but relatively sparse representation in Africa and South America.
Microsoft’s building footprints account for 0.43 billion buildings across multiple continents, and serve as a key supplement to OSM in Europe.
CLSM~(\cite{Shi2024GlobalBuildingFootprints}) and our own generated building polygons primarily address the gaps in Asia, contributing 0.07 and 0.14 billion buildings, respectively.

When comparing building area contributions, the proportion from Open Buildings decreases across continents, suggesting that its average building size is smaller.
In contrast, our generated building polygons contribute more area relative to their building count, likely due to their derivation from lower-resolution satellite imagery, which can lead to the merging of adjacent buildings into single polygon instances.
Despite this limitation, our building polygons significantly complement existing datasets, enabling a more complete and globally consistent coverage of building footprints.

Figure \ref{fig:contribution_sample} presents example polygons generated through the quality-guided polygon fusion process described in Section \ref{sec:quality_guided}.
By integrating additional building footprint datasets, the fusion process substantially enhances the overall completeness and coverage of the final building polygons.

% \subsection{Visual Analyses of Quality-guided Polygon Fusion}

%% The following commands are for the statements about the availability of data sets and/or software code corresponding to the manuscript.
%% It is strongly recommended to make use of these sections in case data sets and/or software code have been part of your research the article is based on.
\noappendix       %% use this to mark the end of the appendix section. Otherwise the figures might be numbered incorrectly (e.g. 10 instead of 1).

%% Regarding figures and tables in appendices, the following two options are possible depending on your general handling of figures and tables in the manuscript environment:

%% Option 1: If you sorted all figures and tables into the sections of the text, please also sort the appendix figures and appendix tables into the respective appendix sections.
%% They will be correctly named automatically.

%% Option 2: If you put all figures after the reference list, please insert appendix tables and figures after the normal tables and figures.
%% To rename them correctly to A1, A2, etc., please add the following commands in front of them:

\appendixfigures  %% needs to be added in front of appendix figures

\appendixtables   %% needs to be added in front of appendix tables

%% Please add \clearpage between each table and/or figure. Further guidelines on figures and tables can be found below.

\authorcontribution{Conceptualization: X.Z.; methodology: X.Z., S.C., F.Z., Y.S.; software: S.C., F.Z., Y.S.; results validation: X.Z., S.C., F.Z.; analysis: X.Z., S.C., F.Z.; data collection: F.Z., S.C., X.Z.; writing, and original draft preparation: X.Z., S.C., F.Z.; paper revision: X.Z., S.C., F.Z., Y.W.; visualization: S.C., F.Z., X.Z., Y.W.; funding acquisition: X.Z.; project administration: X.Z.; resources: X.Z.} %% this section is mandatory

\competinginterests{none} %% this section is mandatory even if you declare that no competing interests are present

\disclaimer{none} %% optional section

\begin{acknowledgements}
Source of the Planet data used in the publication: Planet Labs Inc. This work is jointly supported by the European Research Council (ERC) under the European Union's Horizon 2020 research and innovation programme (grant agreement No. [ERC-2016-StG-714087], Acronym: \textit{So2Sat}), by the German Federal Ministry for Economic Affairs and Climate Action in the framework of the "national center of excellence ML4Earth" (grant number: 50EE2201C), by the German Research Foundation (DFG GZ: ZH 498/18-1; Project number: 519016653), by the Excellence Strategy of the Federal Government and the Länder through the TUM Innovation Network EarthCare, by the European Union (project number: 101091374) and by Munich Center for Machine Learning.
\end{acknowledgements}

%% REFERENCES

%% The reference list is compiled as follows:

%\begin{thebibliography}{}
%\bibitem[AUTHOR(YEAR)]{LABEL1}
%\end{thebibliography}

%% Since the Copernicus LaTeX package includes the BibTeX style file copernicus.bst,
%% authors experienced with BibTeX only have to include the following two lines:
%%
\bibliographystyle{copernicus}
\bibliography{references}

\end{document}